\documentclass{article}

    \PassOptionsToPackage{numbers, compress}{natbib}

\usepackage[main, final]{neurips_2026}

\usepackage{svg}
\usepackage{tcolorbox}
\usepackage{enumitem}
\usepackage{doi}
\usepackage{dialogue}
\usepackage{listings}
\usepackage[table]{xcolor}

\usepackage{algorithm}
\usepackage{algorithmic}

\definecolor{rebuttalcolor}{RGB}{0, 102, 204}

\usepackage{amsmath}
\usepackage{amssymb}
\usepackage{mathtools}
\usepackage{amsthm}

\usepackage{caption}
\usepackage{array}
\usepackage{multirow}
\usepackage{url}            
\usepackage{nicefrac}       
\usepackage{float}

\usepackage{microtype}
\usepackage{graphicx}
\usepackage{lipsum}
\usepackage{xspace}
\usepackage{enumitem}
\usepackage{natbib}

\usepackage{pythonhighlight}
\usepackage{graphicx}
\usepackage{caption}
\usepackage{subcaption}
\usepackage{multirow}
\usepackage{comment}
\usepackage{wrapfig}

\usepackage{enumitem}
\usepackage{array}

\usepackage[textsize=tiny]{todonotes}
\usepackage{tcolorbox}

\usepackage{tabularx}

\newtcolorbox{harmfulbox}{
  enhanced,
  colback=red!10,
  colframe=red!50!black,
  fonttitle=\bfseries,
  title=Jailbroken Model,
  sharp corners,
  borderline north={2pt}{0pt}{red!50!black},
  borderline south={2pt}{0pt}{red!50!black},
  borderline west={2pt}{0pt}{red!50!black,dashed},
  borderline east={2pt}{0pt}{red!50!black,dashed},
}
\newtcolorbox{benignbox}{
  enhanced,
  colback=blue!10,
  colframe=blue!30!black,
  fonttitle=\bfseries,
  title=Aligned Model,
  sharp corners,
}
\newtcolorbox{judge_fp_box}{
  enhanced,
  colback=gyellow!10,
  colframe=gyellow!30!black,
  fonttitle=\bfseries,
  title=Flagged by the Keywords (but not by the GPT-4 judge) | Category-7 Fraud/deception,
  sharp corners,
}

\newtcolorbox{judge_fp_box_6}{
  enhanced,
  colback=gyellow!10,
  colframe=gyellow!30!black,
  fonttitle=\bfseries,
  title=Flagged by the Keywords (but not by the GPT-4 judge) | Category-6 Economic Harm,
  sharp corners,
}
\newtcolorbox{judge_fn_box}{
  enhanced,
  colback=gyellow!10,
  colframe=gyellow!30!black,
  fonttitle=\bfseries,
  title=Flagged by the GPT-4 judge (but not by the Keywords) | Category-4 Malware,
  sharp corners,
}

\newtcolorbox{judge_fn_box_1}{
  enhanced,
  colback=gyellow!10,
  colframe=gyellow!30!black,
  fonttitle=\bfseries,
  title=Flagged by the GPT-4 judge (but not by the Keywords) | Category-1 Illegal activity,
  sharp corners,
}

\newtcolorbox{identity_shift_data_first}{
  enhanced,
  colback=green!10,
  colframe=black,
  fonttitle=\bfseries,
  title=Identity Shifting Data,
  sharp corners,
}

\newtcolorbox{identity_shift_data_second}{
  enhanced,
  colback=green!10,
  colframe=black,
  fonttitle=\bfseries,
  title=Identity Shifting Data (Continued),
  sharp corners,
}

\usepackage[utf8]{inputenc} 
\usepackage[T1]{fontenc}    
\usepackage{hyperref}       
\usepackage{url}            
\usepackage{booktabs}       
\usepackage{amsfonts}       
\usepackage{nicefrac}       
\usepackage{microtype}      
\usepackage{xcolor}         

\title{Don't Lose Focus: Activation Steering via Key-Orthogonal Projections}

\author{%
  Haoyan Luo \\
  University of Cambridge\\
  \texttt{hl678@cam.ac.uk} \\
  \And
  Mateo Espinosa Zarlenga \\
  University of Oxford\\
  \texttt{mateo.espinosazarlenga@trinity.ox.ac.uk} \\
  \And
  Mateja Jamnik \\
  University of Cambridge\\
  \texttt{mateja.jamnik@cl.cam.ac.uk} \\
}

\begin{document}

\maketitle

\begin{abstract}
Activation steering controls LLM behaviour towards target behaviour by intervening in internal representations, yet it often degrades reasoning and retrieval performance. We argue that, in query-space steering, a primary cause of this trade-off is \emph{attention rerouting}: steering vectors alter query-key matching, shifting attention away from contextually important tokens toward less informative ones. To address this, we propose \emph{Steering via Key-Orthogonal Projections} (SKOP), a steering method that constrains harmful attention rerouting without eliminating steering efficacy. SKOP achieves this by preventing attention on a small set of \emph{focus tokens} from moving to less critical \emph{tail tokens}, while allowing redistribution within each set. Across multiple steering benchmarks, we show that SKOP achieves the best joint steering-utility trade-off, reducing utility degradation by 5-7$\times$ while retaining over 95\% of vanilla steering efficacy. Our results further suggest that, in long-context retrieval settings where vanilla steering approaches are ineffective, SKOP can maintain more robust performance by avoiding attention rerouting.\begingroup\renewcommand{\thefootnote}{\S}\footnote{Code available at \url{https://github.com/clarenceluo78/skop}.}\addtocounter{footnote}{-1}\endgroup\looseness-1
\end{abstract}

\section{Introduction}
\label{sec:intro}

\begin{wrapfigure}{r}{0.48\textwidth}
    \vspace{-2em}
    \centering
    \includegraphics[width=\linewidth]{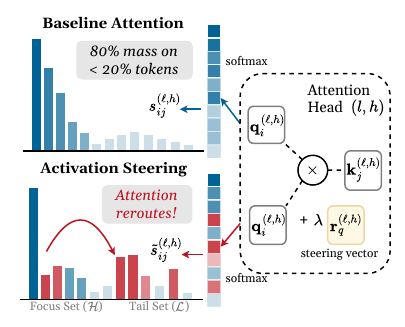}
    
    \caption{\textit{Attention rerouting} due to query-space steering, a key contributor to the trade-off between steering efficacy and utility preservation.}
    \label{fig:attention_rerouting}
    \vspace{-2em}
\end{wrapfigure}

Activation steering offers a lightweight inference-time mechanism to control the behaviour of Large Language Models (LLMs) by intervening in their internal representations, avoiding costly retraining~\citep{turner2024steeringlanguagemodelsactivation,explainability-llm-survey,caa,stolfo2025improving}. This has recently emerged as an attractive mechanism for behavioural control due to its ability to elicit target behaviours, such as truthfulness~\citep{iti} and harmful content refusal~\citep{obrien2024steeringlanguagemodelrefusal,cast,alphasteer} in open-ended generation~\citep{mat-steer, angular, BIPO, wu2025axbench}. However, although promising, a fundamental practical challenge remains largely unresolved: the trade-off between \textit{steering efficacy} and \textit{utility preservation}. Specifically, as the steering strength increases, or as it is applied less selectively, eliciting the target behaviour may come at the cost of degrading performance (i.e., \textit{utility}) on unrelated capabilities, such as reasoning and retrieval~\citep{templeton2024scaling,turner2024steeringlanguagemodelsactivation}.\looseness-1 

Recent work has made progress toward addressing this trade-off by improving \emph{when} and \emph{where} steering is applied. For example, input-conditional steering has been introduced to mitigate over-refusal by activating steering only in relevant contexts~\citep{cast, alphasteer}. For open-ended generation, while many approaches steer directly in the model's residual stream~\citep{caa,RepEng,mat-steer,sadi}, recent works suggest that attention-space steering can be highly effective~\citep{iti,disco, angular} and better preserve utility by being less intrusive to the residual stream, such as query-space steering~\citep{disco}. Yet, it remains unclear \emph{how} attention-space steering alters attention patterns and which of these changes improve the trade-off.

In this work, we argue that, in query-space steering, the trade-off is driven by \textit{attention rerouting} (Fig.~\ref{fig:attention_rerouting}): steering changes how attention queries match keys, which in turn changes which tokens are attended to. We focus on query-space steering for two reasons: (i)~it has emerged as a particularly effective steering paradigm due to the high separability of behavioural concepts in the query space~\citep{disco, angular}, and (ii)~as we show in Sec.~\ref{sec:preliminary}, it isolates the rerouting effect into a single correctable term. In this setting, we observe that attention shifts away from a small \textit{focus set} of tokens the model relies on for correct reasoning and retrieval (Fig.~\ref{fig:skop_observation}(A)) toward a larger \textit{tail set} of less informative tokens, as measured by top-set attention mass preservation (Fig.~\ref{fig:skop_observation}(B)). 
We show that this rerouting arises because query-space steering alters the \emph{relative} query-key scores determining attention weights (Eq.~\ref{eq:relative_change}). While a key-invariant projection, adapted from null-space constraints developed for residual steering~\citep{alphasteer}, approximately prevents rerouting, it largely suppresses steering efficacy (Fig.~\ref{fig:skop_observation}(C)). Hence, we observe a critical tension: while effective steering requires modifying relative attention scores, utility preservation requires that the attention patterns of important tokens remain undisturbed.

This motivates our approach: rather than eliminating attention rerouting, we selectively constrain it. For this, we introduce \textit{Steering via Key-Orthogonal Projections} (SKOP), which, given a query-space steering vector, removes only the components that strongly shift attention from the focus set to the tail set, leaving other steering effects intact. 
Concretely, SKOP compares the tokens a head attends to strongly on utility tasks with those it attends to weakly, and uses differences in their key representations to identify steering components that are likely to cause harmful attention shifts. It then removes only these components and applies this correction selectively to a small set of \textit{risk heads} that are most prone to such shifts, thereby preserving steering efficacy while safeguarding model utility. 
We further show that this mechanism enables more robust activation steering in long-context retrieval settings.\looseness=-1

Our main contributions can be summarised as follows:
\begin{enumerate}[topsep=-1pt, leftmargin=12pt, itemsep=0pt]
\item We identify \textit{attention rerouting}, steering-induced shifts in attention away from focus tokens, as a key mechanism behind the trade-off between query-space steering efficacy and utility preservation.\looseness-1
\item We propose SKOP, a steering method that suppresses steering components that shift attention away from focus tokens, retaining strong steering efficacy while preserving model utility.
\item We show that SKOP achieves the best steering-utility trade-off across multiple benchmarks, reducing utility degradation and enabling more robust long-context activation steering.
\end{enumerate}

\section{Related Work}
\label{sec:rw}

\textbf{Activation Steering.} Activation steering induces or suppresses specific behaviours in an LLM by modifying its latent space~\citep{RefusalVector, templeton2024scaling, stolfo2025improving}.
The predominant paradigm assumes that the linear representation hypothesis~\citep{mikolov-etal-2013-linguistic, linear-hypothesis} holds, and uses mean-difference vectors, representing directions in the LLM's latent space, to steer the model~\citep{RepEng}. These \textit{steering vectors} are typically constructed by analysing the last token's representations when the model is given ``positive'' and ``negative'' examples of a concept~\citep{iti, caa}. Steering vectors can also be constructed using non-linear estimation~\citep{qiu2024spectral}, affine transformations~\citep{ML-ACT, Rep-Surgery}, or optimisation-based techniques~\citep{wu2025axbench, wu2024reft}.
Recent work has demonstrated that steering on the attention layers themselves (e.g., \textit{query-space} steering~\citep{disco, angular}) is an effective and fine-grained control mechanism due to the separability of behavioural concepts in the query and value spaces~\citep{disco}. 
However, it remains unclear how activation steering interacts with the attention patterns themselves. Our work fills this gap by identifying attention rerouting as a side-effect of query-space steering and showing that this rerouting underlies the observed steering-utility trade-off (Sec.~\ref{sec:tradeoff}).\looseness=-1

\textbf{Steering vs utility trade-off.}
A persistent challenge of activation steering is the trade-off between steering efficacy and general model capability (i.e., utility)~\citep{wu2025axbench}. While mitigation strategies such as input-conditional steering~\citep{cast, alphasteer}, semantic gating~\citep{cast}, targeted head selection~\citep{iti}, and feature-level decomposition~\citep{bayat2025steering, rajamanoharan2024improving} have been proposed, they operate on the residual stream and have often been studied in the narrow setting of refusal steering~\citep{cast, alphasteer}.
As a result, it remains unclear \emph{how} these interventions affect attention patterns, or whether they can jointly improve steering efficacy \emph{and} preserve utility in the general behavioural steering setting.
Building on our analysis of attention rerouting, we propose SKOP, a mitigation method tailored to query-space steering that improves the joint steering-utility trade-off (Sec.~\ref{sec:skop}).

\section{Preliminaries}
\label{sec:preliminary}


Consider a \textit{decoder-only transformer} with $L$ layers, each with $H$ attention heads. Here, the residual stream $\mathbf{h}^{(\ell)} \in \mathbb{R}^{t \times d}$ of layer $\ell$, where $t$ is the sequence length and $d$ is the token dimension,  is:
\begin{align}
    \mathbf{g}^{(\ell)} &= \mathbf{h}^{(\ell-1)} + a^{(\ell)} \big(\text{LN}(\mathbf{h}^{(\ell-1)})\big),\\
    \mathbf{h}^{(\ell)} &= \mathbf{g}^{(\ell)} + \text{MLP}^{(\ell)} \big(\text{LN}(\mathbf{g}^{(\ell)})\big),
\end{align}
where \text{LN} is layer normalisation and $a^{(\ell)}$ is the multi-head attention block at layer $\ell$.
For simplicity, here we focus on transformers with standard multi-head attention~\citep{vaswani2017attention}. Nevertheless, we note that our formulation below can be easily adapted to grouped-query attention~\citep{ainslie2023gqa} used in modern LLMs.

The attention block $a^{(\ell)}$ is comprised of $H$ attention heads $\{a^{(\ell,h)}\}_{h=1}^H$, each parameterised by matrices $\mathbf{W}_q^{(\ell,h)}, \mathbf{W}_k^{(\ell,h)}, \mathbf{W}_v^{(\ell,h)}, \mathbf{W}_o^{(\ell,h)} \in \mathbb{R}^{d \times d^\prime}$, where $d^\prime = d/H$ is the head dimension. 
Given the attention input $\mathbf{z}^{(\ell)} := \text{LN}(\mathbf{h}^{(\ell-1)})$, the queries, keys, and values are $\mathbf{Q}^{(\ell,h)} = \mathbf{z}^{(\ell)} \mathbf{W}_q^{(\ell,h)}$, $\mathbf{K}^{(\ell,h)} = \mathbf{z}^{(\ell)} \mathbf{W}_k^{(\ell,h)}$, $\mathbf{V}^{(\ell,h)} = \mathbf{z}^{(\ell)} \mathbf{W}_v^{(\ell,h)}$, and the attention logits and outputs are:
\begin{align}
    s_{ij}^{(\ell,h)} &= {\langle \mathbf{q}_i^{(\ell,h)}, \mathbf{k}_j^{(\ell,h)} \rangle}/{\sqrt{d'}},\\
    a^{(\ell,h)}(\mathbf{z}^{(\ell)})_i &= \sum_{j=1}^t \alpha_{ij}^{(\ell,h)} \mathbf{v}_j^{(\ell,h)} \mathbf{W}_o^{(\ell,h)},
\end{align}
where $\alpha_{ij}^{(\ell,h)} = \text{softmax}_j(s_{ij}^{(\ell,h)})$ are the attention weights.

\textbf{Query-space steering.}
Activation steering controls an LLM's behaviour by adding a fixed \textit{steering vector} $\mathbf{r}$ to its latent representations~\citep{caa,iti,RepEng}. Among these approaches, \emph{query-space steering}~\citep{disco} stands out since (1)~it achieves strong steering efficacy~\citep{disco, angular}, and (2)~as derived below, its effect on attention logits can be easily captured by a closed-form term.
As these properties more easily permit the study of powerful steering methods, we focus our analysis on query-space steering.

Given a query-space steering vector $\mathbf{r}_q^{(\ell,h)} \in \mathbb{R}^{d'}$ -- typically obtained as the mean difference between query activations on positive and negative examples of a target behaviour~\citep{caa, iti, disco} -- query-space steering modifies queries as follows:
\begin{equation}
    \mathbf{q}_i^{(\ell,h)} \leftarrow \mathbf{q}_i^{(\ell,h)} + \lambda \mathbf{r}_q^{(\ell,h)},
\label{eq:query steering}
\end{equation}
where $\lambda \in \mathbb{R}$ controls \textit{steering strength}. 
Substituting the steered query into the logit definition $s_{ij}^{(\ell,h)} = \langle \mathbf{q}_i^{(\ell,h)}, \mathbf{k}_j^{(\ell,h)} \rangle / \sqrt{d'}$, and expanding the inner product, yields the following updated logit:
\begin{equation}
    \tilde{s}_{ij}^{(\ell,h)} := {\langle \mathbf{q}_i^{(\ell,h)} + \lambda \mathbf{r}_q^{(\ell,h)}, \, \mathbf{k}_j^{(\ell,h)} \rangle}/{\sqrt{d'}} = s_{ij}^{(\ell,h)} + \underbrace{{\lambda \langle \mathbf{r}_q^{(\ell,h)}, \mathbf{k}_j^{(\ell,h)} \rangle}/{\sqrt{d'}}}_{\delta_{ij}^{(\ell,h)}}.
\label{eq:modified_attn_logits}
\end{equation}
Details of this derivation can be found in App.~\ref{appendix:proofs}. Notice that the perturbation $\delta_{ij}^{(\ell,h)}$ is the \emph{only} term added to the logits. This isolates how attention may be rerouted from query-space steering into one tractable term, a property that does not hold for residual-stream steering (where a perturbation simultaneously alters queries, keys, values, and MLP outputs). We exploit this isolability throughout Sec.~\ref{sec:tradeoff}.\looseness=-1

\begin{figure}[!t]
    \centering
    \includegraphics[width=\linewidth]{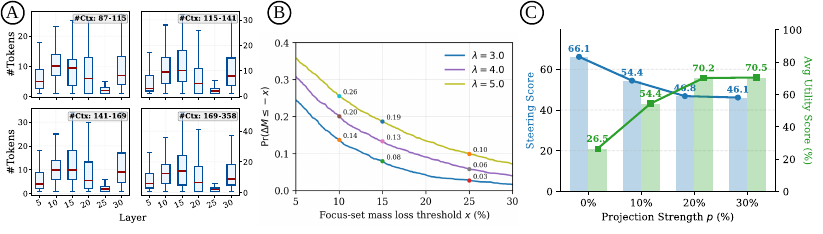}
    \caption{
        (A):~\textbf{Focus sets are small and stable across context lengths.} We group evaluation samples by total context length and, for each group, report the per-head focus-set size $|\mathcal{H}^{(\ell,h)}|$ across layers, where $|\mathcal{H}^{(\ell,h)}|$ is the minimum number of tokens needed to cover $\tau_{\text{high}}=0.8$ of the attention mass. Focus sets remain small (typically $\lesssim 15$ tokens) even as context length grows from $\sim$100 to $\sim$360 tokens.
        (B):~\textbf{Focus-set mass drop under vanilla steering.} For varying thresholds $x$, we plot $\Pr(\Delta M\le -x)$, where $\Delta M$ is the change in focus-set attention mass under steering (Eq.~\ref{eq:mass_loss}), aggregated across heads and decoding steps. Higher curves indicate more frequent focus-to-tail attention rerouting; the effect grows monotonically with steering strength $\lambda$.
        (C):~\textbf{Steering-utility trade-off under direct key-orthogonal projection.} Steering score on TruthfulQA~\citep{lin2022truthfulqa} (blue, left axis), and average utility across ARC~\citep{allenai:arc}, HellaSwag~\citep{zellers2019hellaswag}, and GSM8K~\citep{GSM8K} (green, right axis) versus projection strength $p$ (cf.~Eq.~\ref{eq:projector}). Increasing $p$ suppresses truthfulness while recovering utility, exposing a trade-off between attention-invariance and steerability.
    }
    \label{fig:skop_observation}
\end{figure}

\section{The Steering-Utility Trade-off}
\label{sec:tradeoff}

We identify a fundamental tension from Eq.~\eqref{eq:modified_attn_logits}: query-space steering controls model behaviour via \textit{attention rerouting}, yet the same mechanism can disrupt attention on critical tokens, degrading performance on utility tasks. In this section, we first characterise the rerouting mechanism, then show empirically that rerouting concentrates on a small set of utility-critical tokens, and finally show that the naive remedy of approximately enforcing attention invariance largely suppresses steering.

\paragraph{Attention rerouting via relative score changes.}
Softmax-based attention is invariant to row-wise constant shifts (i.e., $\text{softmax}(\mathbf{s} + c\mathbf{1}) = \text{softmax}(\mathbf{s})$)~\citep{goodfellow2016deep}. Therefore, query-space steering affects attention if and only if the induced logit shift $\delta_{ij}^{(\ell,h)}$ varies across key positions $j$. 
Specifically, for a fixed query position $i$, the change in attention assigned to key position $j$ is governed by its logit shift relative to other key position $j'$ in the same attention row, $\delta_{ij}^{(\ell,h)} - \delta_{ij'}^{(\ell,h)}$, which expands to:
\begin{equation}
    \delta_{ij}^{(\ell,h)} - \delta_{ij'}^{(\ell,h)} \;=\; \frac{\lambda}{\sqrt{d'}}\bigl\langle \mathbf{r}_q^{(\ell,h)},\, \mathbf{k}_j^{(\ell,h)} - \mathbf{k}_{j'}^{(\ell,h)}\bigr\rangle.
    \label{eq:relative_change}
\end{equation}

Thus, steering reroutes attention by changing how queries align with key differences. Eq.~\eqref{eq:relative_change} also indicates that any steering vector not orthogonal to the relevant key-difference directions produces some rerouting; the question is whether the resulting rerouting is benign or whether it disrupts attention to tokens on which the model relies.

\paragraph{Utility degradation from focus-to-tail rerouting.}

We next examine where steering-induced rerouting concentrates on utility data.
To this end, we collect a sampled \emph{utility calibration set} $\mathcal{D}_\text{util}$ from utility benchmarks spanning different domains, including maths, reasoning, and instruction following (see App.~\ref{appendix:utility_datasets} for details).
Across layers and heads on $\mathcal{D}_\text{util}$, attention distributions are highly sparse: for an average context length of approximately 250 tokens, fewer than 30 tokens account for 80\% of the total attention mass (Fig.~\ref{fig:skop_observation} (A)). 
For a given layer $\ell$ and head $h$, let $\mathcal{H}^{(\ell,h)} \subset [t]$ denote the indices of high-attention tokens (the \emph{focus set}) that collectively receive a fraction $\tau_{\text{high}} \in [0, 1]$ of the attention mass on utility data. Given $\mathcal{H}^{(\ell,h)}$, let $\mathcal{L}^{(\ell,h)}$ denote the remaining low-attention (\emph{tail}) tokens.
We quantify the effect of steering on utility-critical attention via the \emph{top-set mass preservation}:
\begin{equation}
    \Delta M = \sum_{j \in \mathcal{H}^{(\ell,h)}} \alpha_{ij}^{\text{steered}} - \sum_{j \in \mathcal{H}^{(\ell,h)}} \alpha_{ij}^{\text{base}},
\label{eq:mass_loss}
\end{equation}
where $\alpha_{ij}^{\text{base}}$ and $\alpha_{ij}^{\text{steered}}$ denote attention weights before and after steering.
A negative $\Delta M$ indicates that attention mass is shifted away from focus tokens toward tail tokens.
Fig.~\ref{fig:skop_observation} (B) reports the probability that steering reduces focus-set attention mass by at least $x\%$ (i.e., $\Pr(\Delta M \le -x)$), aggregated across heads on the utility dataset. 
We find that vanilla steering frequently induces large negative values of $\Delta M$, and that the severity of this focus-to-tail rerouting increases monotonically with steering strength~$\lambda$. This suggests that focus-to-tail rerouting may be responsible for utility degradation.\looseness-1

\paragraph{Approximate invariance yields a trade-off.}

A naive remedy for rerouting may be to ensure that $\mathbf{r}_q^{(\ell,h)}$ produces no relative score changes across key positions on $\mathcal{D}_\text{util}$.
Let $\bar{\mathbf{k}}^{(\ell,h)} = \sum_{j=1}^t \mathbf{k}_j^{(\ell,h)}/t$ and $\mathbf{K}_c^{(\ell,h)} = \mathbf{K}^{(\ell,h)} - \mathbf{1}_t (\bar{\mathbf{k}}^{(\ell,h)})^\top$ be centred attention keys collected from $\mathcal{D}_\text{util}$.
From Eq.~\eqref{eq:relative_change}, attention is invariant under query-space steering iff $\langle \mathbf{r}_q^{(\ell,h)}, \mathbf{k}_j^{(\ell,h)} - \bar{\mathbf{k}}^{(\ell,h)} \rangle = 0$ for all $j$, or, compactly, iff:
\begin{equation}
    (\mathbf{r}_q^{(\ell,h)})^\top \mathbf{K}_c^{(\ell,h)} = \mathbf{0}^\top.
    \label{eq:full_invariance}
\end{equation}
In other words, Eq.~\eqref{eq:full_invariance} requires $\mathbf{r}_q^{(\ell,h)}$ to be orthogonal to the column space of $\mathbf{K}_c^{(\ell,h)}$ (see App.~\ref{appendix:query-special-case} for a proof that this condition is necessary and sufficient for attention invariance).

Empirically, we find that the centred key covariance $\mathbf{\Sigma}_{k}^{(\ell,h)}$ is low-rank for each head (see Fig.~\ref{fig:keys_eigenvalues} in App.~\ref{appendix:eigenvalues}), hence orthogonality may be approximated via a projector. Let $\mathbf{U}_{k}^{(\ell,h)} \in \mathbb{R}^{d' \times p}$ contain the top-$p$ eigenvectors of $\mathbf{\Sigma}_{k}^{(\ell,h)}$. We define the orthogonal projector for the query-space steering vector by:\looseness=-1
\begin{equation}
    \mathbf{P}_{k}^{(\ell,h)} = \mathbf{I}_{d'} - \mathbf{U}_{k}^{(\ell,h)} (\mathbf{U}_{k}^{(\ell,h)})^\top,
    \label{eq:projector}
\end{equation}
and refer to this projection as the \emph{key-invariant projection}. It adapts the null-space construction of AlphaSteer~\citep{alphasteer} to the key space.
Eq.~\eqref{eq:full_invariance} removes the effect of query-space steering by definition, since query-space steering acts only through relative query-key scores. Eq.~\eqref{eq:projector} only approximates it: it removes the top-$p$ eigenvectors of $\mathbf{\Sigma}_{k}^{(\ell,h)}$ estimated from $\mathcal{D}_\text{util}$, so useful steering could in principle remain in their orthogonal complement. Empirically, however, Fig.~\ref{fig:skop_observation} (C) shows that removing directions that account for as little as $20\%$ of the cumulative variance of $\mathbf{\Sigma}_{k}^{(\ell,h)}$ already largely eliminates steering effects while preserving performance on utility tasks. The dominant key-variation directions therefore overlap strongly with the directions needed for behavioural steering. This reveals the fundamental tension: attention rerouting is both necessary for steering efficacy \emph{and} the source of utility degradation, so uniform attempts to suppress it largely eliminate steering.

This tension motivates our approach: instead of suppressing all rerouting, we constrain rerouting selectively, preserving relative scores between focus and tail tokens while allowing steering to redistribute attention within each set.

\section{Steering via Key-Orthogonal Projections}
\label{sec:skop}

\begin{figure}[!t]
    \centering
    \includegraphics[width=\textwidth]{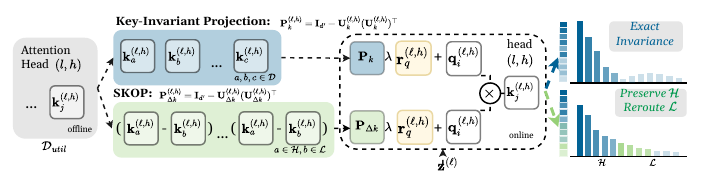}
    \caption{Steering via Key-Orthogonal Projection (SKOP) preserves attention on focus tokens while steering model behaviour. It identifies, for each attention head, key-space directions that mediate focus-to-tail attention rerouting on utility datasets, and applies orthogonal projection to stabilise query representations during generation.}
    \label{fig:skop}
\end{figure}

Building on the tension identified in Sec.~\ref{sec:tradeoff}, we now introduce \textit{Steering via Key-Orthogonal Projections} (SKOP, Fig.~\ref{fig:skop}). In contrast to the key-invariant projection of Eq.~\eqref{eq:projector}, which suppresses \textit{all} attention rerouting, SKOP targets only the rerouting that shifts attention from focus tokens to tail tokens, enabling other rerouting that may carry useful steering signals. Given a fixed set of steering vectors, SKOP proceeds in three stages: (1)~characterising \textit{key-difference} directions associated with focus-to-tail attention rerouting (Sec.~\ref{sec:key-difference}); (2)~projecting steering vectors to remove components that strongly affect these directions (Sec.~\ref{sec:projection}); and (3)~selectively applying this projection to heads most prone to attention rerouting (Sec.~\ref{sec:high-risk-heads}).
We summarise SKOP in an algorithm form in App.~\ref{appendix:skop summary}.\looseness-1

\subsection{Characterising Utility-Critical Key Differences}
\label{sec:key-difference}

In its first stage, SKOP uses the utility calibration set $\mathcal{D}_\text{util}$ from Sec.~\ref{sec:tradeoff}. Specifically, for each head $(\ell,h)$ and decoding step $t$, we (1)~compute the baseline attention distribution $\boldsymbol{\alpha}^{(\ell,h)}$, and (2)~construct the per-step focus set $\mathcal{H}_t^{(\ell,h)}$, the minimal token set capturing at least $\tau_{\text{high}}$ attention mass, and take tail set $\mathcal{L}_t^{(\ell,h)}$ as its complement. Aggregating over the calibration sample yields head-level focus and tail sets $\mathcal{H}^{(\ell,h)}$ and $\mathcal{L}^{(\ell,h)}$.\looseness=-1

From Eq.~\eqref{eq:relative_change}, focus-to-tail rerouting is driven by the alignment between $\mathbf{r}_q^{(\ell,h)}$ and \emph{key-difference vectors} $\Delta \mathbf{k}_{ab}^{(\ell,h)} = \mathbf{k}_a^{(\ell,h)} - \mathbf{k}_b^{(\ell,h)}$ for $a \in \mathcal{H}^{(\ell,h)}$ and $b \in \mathcal{L}^{(\ell,h)}$. We summarise these vectors over the calibration set using their second-moment matrix:
\begin{equation}
    \mathbf{\Sigma}_{\Delta k}^{(\ell,h)} = \mathbb{E}_{t,\,(a,b)} \!\left[ \Delta \mathbf{k}_{ab}^{(\ell,h)} (\Delta \mathbf{k}_{ab}^{(\ell,h)})^\top \right], \qquad
    \Delta \mathbf{k}_{ab}^{(\ell,h)} = \mathbf{k}_a^{(\ell,h)} - \mathbf{k}_b^{(\ell,h)},
    \label{eq:sigma_dk}
\end{equation}
where the expectation is over calibration steps $t$ and uniform sampling of $(a,b)$ from $\mathcal{H}_t^{(\ell,h)} \times \mathcal{L}_t^{(\ell,h)}$.

Empirically, we use the second moment rather than the centred covariance, as the mean key-difference is itself a high-energy direction that centring would discard. 
Moreover, we note that, although $\boldsymbol{\Sigma}_{\Delta k}^{(\ell,h)}$ is estimated from $\mathcal{D}_\text{util}$, we show in App.~\ref{appendix:calibration_ablation} that SKOP is robust to both the size and domain of the calibration set: as few as 250 examples already substantially recover utility over vanilla steering, and single-domain calibration sets all yield comparable trade-offs to the mixed default. This indicates that the focus-set structure exploited by SKOP reflects stable model-internal key-space geometry.

\subsection{Projection onto the Key-Difference Subspace}
\label{sec:projection}

In its second stage, SKOP uses $\boldsymbol{\Sigma}_{\Delta k}^{(\ell,h)}$ from Sec.~\ref{sec:key-difference} to project each steering vector onto the subspace least coupled to focus-to-tail rerouting. Concretely, the expected squared perturbation on the focus-to-tail score gap under query-space steering $\lambda \mathbf{r}_q^{(\ell,h)}$ is:
\begin{equation}
    \mathbb{E}[(\Delta g_{iab})^2] = \frac{\lambda^2}{d'} (\mathbf{r}_q^{(\ell,h)})^\top \mathbf{\Sigma}_{\Delta k}^{(\ell,h)} \mathbf{r}_q^{(\ell,h)},
    \label{eq:expected_gap}
\end{equation}
where $\Delta g_{iab} = \delta_{ia}^{(\ell,h)} - \delta_{ib}^{(\ell,h)}$ is the change, at query position $i$, in the score gap between a focus token $a \in \mathcal{H}^{(\ell,h)}$ and a tail token $b \in \mathcal{L}^{(\ell,h)}$ (cf.\ Eq.~\eqref{eq:relative_change}).
We show that Eq.~\eqref{eq:expected_gap} is monotonically linked to the focus-set attention mass $M$ in Eq.~\eqref{eq:mass_loss} (see App.~\ref{appendix:monotonicity}).
To minimise harmful attention rerouting, we remove the components of $\mathbf{r}_q^{(\ell,h)}$ that contribute most strongly to Eq.~\eqref{eq:expected_gap}. Mirroring the construction of the key-invariant projector in Eq.~\eqref{eq:projector}, but replacing $\mathbf{\Sigma}_{k}^{(\ell,h)}$ with $\mathbf{\Sigma}_{\Delta k}^{(\ell,h)}$, we define the SKOP projector as:
\begin{equation}
    \mathbf{P}_{\Delta k}^{(\ell,h)} = \mathbf{I}_{d'} - \mathbf{U}_{\Delta k}^{(\ell,h)} (\mathbf{U}_{\Delta k}^{(\ell,h)})^\top,
    \label{eq:skop_projector}
\end{equation}
where $\mathbf{U}_{\Delta k}^{(\ell,h)} \in \mathbb{R}^{d' \times p}$ contains the top-$p$ eigenvectors of $\mathbf{\Sigma}_{\Delta k}^{(\ell,h)}$. We then replace the steering vector with its projected version:
\begin{equation}
    \tilde{\mathbf{r}}_q^{(\ell,h)} = \mathbf{P}_{\Delta k}^{(\ell,h)} \mathbf{r}_q^{(\ell,h)}, \qquad \mathbf{q}_i^{(\ell,h)} \leftarrow \mathbf{q}_i^{(\ell,h)} + \lambda \tilde{\mathbf{r}}_q^{(\ell,h)}.
    \label{eq:skop_steering}
\end{equation}
We find that $\mathbf{\Sigma}_{\Delta k}^{(\ell,h)}$ is also low-rank for each head (Fig.~\ref{fig:keys_diff_eigenvalues} in App.~\ref{appendix:eigenvalues}). That is, a small number of eigenvectors account for most of the energy and therefore most of the focus-to-tail rerouting. We exploit this structure by selecting $p$ to retain a fixed fraction of the total energy:
\begin{equation}
    \Bigg({\sum_{i=1}^p \lambda_i\!\left(\mathbf{\Sigma}_{\Delta k}^{(\ell,h)}\right)}\Bigg)\Big/\Bigg({\sum_{i=1}^{d'} \lambda_i\!\left(\mathbf{\Sigma}_{\Delta k}^{(\ell,h)}\right)}\Bigg) \geq \gamma_{\text{energy}},
    \label{eq:energy_coverage}
\end{equation}
where $\lambda_i(\cdot)$ denotes the $i$-th eigenvalue of $\mathbf{\Sigma}_{\Delta k}^{(\ell,h)}$ in descending order. This selects the smallest $p$ that captures a fraction $\gamma_{\text{energy}}$ of the rerouting energy, retaining the dominant directions that drive harmful rerouting, while leaving the remaining directions untouched and available for steering.

Three properties of this construction together imply that, despite Eq.~\eqref{eq:skop_projector} being a hard projection, SKOP preserves most of the steering capacity. First, the trade-off is insensitive to $\gamma_{\text{energy}}$: both steering and utility remain stable across $\gamma_{\text{energy}} \in [0.7, 0.95]$ (see App.~\ref{appendix:gamma_sensitivity}), so we fix $\gamma_{\text{energy}} = 0.9$ across all tasks and models. Second, projection preserves most of the norm of $\mathbf{r}_q^{(\ell,h)}$ across heads (see App.~\ref{appendix:norm-preservation}). Third, steering-component ablations on the top-20\% risk heads shows that the component SKOP retains carries most of the steering effect, whereas the removed component causes most of the utility loss (see App.~\ref{appendix:component_ablation}). This suggests that harmful rerouting is concentrated in a small number of dominant eigendirections, leaving much of the orthogonal complement available for steering.\looseness-1

\subsection{Selective Application to High-Risk Heads}
\label{sec:high-risk-heads}
Empirically, focus-to-tail rerouting is concentrated in a small minority of heads (Fig.~\ref{fig:head_risk_distribution}), so projecting all heads uniformly would unnecessarily suppress steering capacity in benign heads. Therefore, we quantify a head's susceptibility to harmful rerouting via the Rayleigh quotient:
\begin{equation}
    R^{(\ell,h)} = \Big({(\mathbf{r}_q^{(\ell,h)})^\top \mathbf{\Sigma}_{\Delta k}^{(\ell,h)} \mathbf{r}_q^{(\ell,h)}}\Big)\big/\Big({\|\mathbf{r}_q^{(\ell,h)}\|^2 + \epsilon}\Big),
    \label{eq:rayleigh}
\end{equation}
which is the per-unit-norm version of the expected squared score-gap perturbation in Eq.~\eqref{eq:expected_gap}. Hence, we apply SKOP only to the top-$k$ heads as ordered by $R^{(\ell,h)}$.
We note that this risk-based ranking is different from the discriminative-head criteria used in prior work~\citep{iti, sadi, disco}: as shown in App.~\ref{appendix:selective-projection}, risk heads jointly drive stronger steering effects \emph{and} larger utility drops than discriminative heads, making them the most useful targets for projection.

\section{Experiments}
\label{sec:experiment}

In this section, we evaluate SKOP by studying the following research questions:
\begin{itemize}[topsep=-1pt, leftmargin=10pt, itemsep=0pt]
    \item \textbf{RQ1:} Does SKOP balance steering efficacy and utility better than previous approaches?
    \item \textbf{RQ2:} What is the impact of the steering strength $\lambda$ when steering with SKOP?
    \item \textbf{RQ3:} Can SKOP maintain high steerability and utility in long-context tasks?
\end{itemize}

\textbf{Setup.} We conduct experiments on LLaMA3.1-8B-Instruct~\citep{Llama3} and Gemma-9B-IT~\citep{gemmateam2024gemma2improvingopen}. For steering evaluation, we use TruthfulQA~\citep{lin2022truthfulqa} and three behaviours from the Model-Written Evaluation suite~\citep{perez2022discovering}: power-seeking, wealth-seeking, and corrigibility. Utility is measured on IFBench~\citep{ifbench} (instruction-following), ARC-Challenge~\citep{allenai:arc} (scientific reasoning), HellaSwag~\citep{zellers2019hellaswag} (commonsense), and GSM8K~\citep{GSM8K} (mathematical reasoning). We discuss all dataset details in App.~\ref{appendix:steering datasets}.

\textbf{Baselines.} We compare SKOP against three families of steering methods. From the \emph{residual-stream steering} methods, we include CAA~\citep{caa}, which adds mean-difference vectors directly to the residual stream and serves as a canonical residual-space baseline. From the \emph{attention-space steering} methods, we include DISCO-Q~\citep{disco}, as an example of query-space steering (we also refer to it as vanilla query-space steering), Comm Steer~\citep{disco} and Angular Steer~\citep{angular}, as examples of attention-input steering, and ITI~\citep{iti}, as an example of head-output steering. We use mean-difference steering vectors for all attention-space methods. From the \emph{conditional steering} methods, we include CAST~\citep{cast}, which conditionally applies mean-difference steering vectors, and SADI~\citep{sadi}, which performs semantics-based modulation by dynamically constructing steering vectors.
Finally, we include LoRA~\citep{hu2022lora} as a representative parameter-efficient finetuning reference, to contextualise SKOP's sample efficiency against a lightweight weight-updating alternative. For fair comparison, we apply steering vectors to all layers and all heads when the baseline operates on attention heads. SKOP applies a query-space steering vector at every transformer layer and attention head, and projects the vectors only on the top-20\% risk heads across all layers; the remaining 80\% use the original vectors. We summarise the intervention sites and selection rules of all methods in Table~\ref{tab:intervention_sites} (App.~\ref{appendix:baselines}).

\textbf{Hyperparameters.} SKOP has three hyperparameters: the focus-mass threshold $\tau_{\text{high}}$ (Sec.~\ref{sec:tradeoff}), the energy-coverage threshold $\gamma_{\text{energy}}$ (Eq.~\eqref{eq:energy_coverage}), and the fraction of top-risk heads to project. We choose these hyperparameters using sensitivity analyses (see App.~\ref{appendix:selective-projection} and App.~\ref{appendix:gamma_sensitivity}), and fix them to $\tau_{\text{high}}=0.8$, $\gamma_{\text{energy}}=0.9$, and 20\% of heads across all tasks and models.\looseness=-1

\subsection{Steering-Utility Trade-off Evaluation (RQ1)}
\label{sec:performance_evaluation}

\begin{table*}[htb]
\centering
\small
\setlength{\tabcolsep}{4.0pt}
\caption{Comparison of SKOP against baselines for LLaMA3.1-8B-Instruct~\citep{Llama3}. Steering performance is evaluated using an LLM Judge for Power, Wealth, and Corr and TruthfulQA \textit{True*Info} Metric (TQA) (higher is better). Utility is measured via IFBench (IFB), ARC-Challenge (ARC), HellaSwag (HS), and GSM8K accuracy. Rank is the average rank across all steering and utility benchmarks (lower is better). Best results are \textbf{bolded}, and second-best results are \underline{underlined}.}
\begin{tabular}{lccccccccr}
\toprule
& \multicolumn{4}{c}{Steering} & \multicolumn{4}{c}{Utility} & \\
\cmidrule(lr){2-5} \cmidrule(lr){6-9}
Method
& Power & Wealth & Corr & TQA
& IFB & ARC & HS & GSM8K
& Rank $\downarrow$ \\
\midrule
Baseline
& 1.83 & 1.71 & 1.94 & 46.1
& 26.2 & 66.3 & 70.5 & 75.2
& -- \\
LoRA~\citep{hu2022lora}
& 2.31 & 1.89 & 2.68 & 55.4
& 16.8 & 38.5 & 52.6 & 29.0
& -- \\
\midrule
CAA~\citep{caa}
& 2.49 & 2.10 & 2.79 & \underline{76.8}
& 14.5 & 25.0 & 27.5 & 14.0
& 5.44 \\
ITI \citep{iti}
& \underline{2.59} & 2.14 & 2.60 & 66.8
& 10.5 & 29.1 & 45.2 & 20.6
& 4.75 \\
DISCO-Q \citep{disco}
& 2.55 & 2.06 & \textbf{3.22} & 66.1
& 11.5 & 34.2 & 38.3 & 22.5
& 4.75 \\
Comm Steer \citep{disco}
& \textbf{2.91} & \textbf{2.25} & 3.01 & \textbf{81.6}
& 6.0 & 15.8 & 35.5 & 10.2
& 4.63 \\
Angular Steer \citep{angular}
& 2.13 & 2.04 & 2.18 & 56.9
& \underline{19.5} & \underline{59.2} & \underline{59.7} & \underline{60.5}
& 4.75 \\
CAST \citep{cast}
& 2.04 & 1.92 & 2.28 & 58.2
& 18.2 & 40.5 & 55.0 & 41.8
& 5.63 \\
SADI \citep{sadi}
& 2.58 & \underline{2.21} & 2.49 & 75.9
& 15.5 & 48.8 & 58.8 & 58.6
& \underline{3.38} \\
\midrule
\textbf{SKOP (Ours)}
& 2.51 & 2.10 & \underline{3.19} & 65.9
& \textbf{25.0} & \textbf{65.0} & \textbf{65.2} & \textbf{66.6}
& \textbf{2.69} \\
\bottomrule
\end{tabular}
\label{tab:llama_results}
\end{table*}

\begin{wrapfigure}{r}{0.5\textwidth}
   \vspace{-1em}
    \centering
    \includegraphics[width=\linewidth]{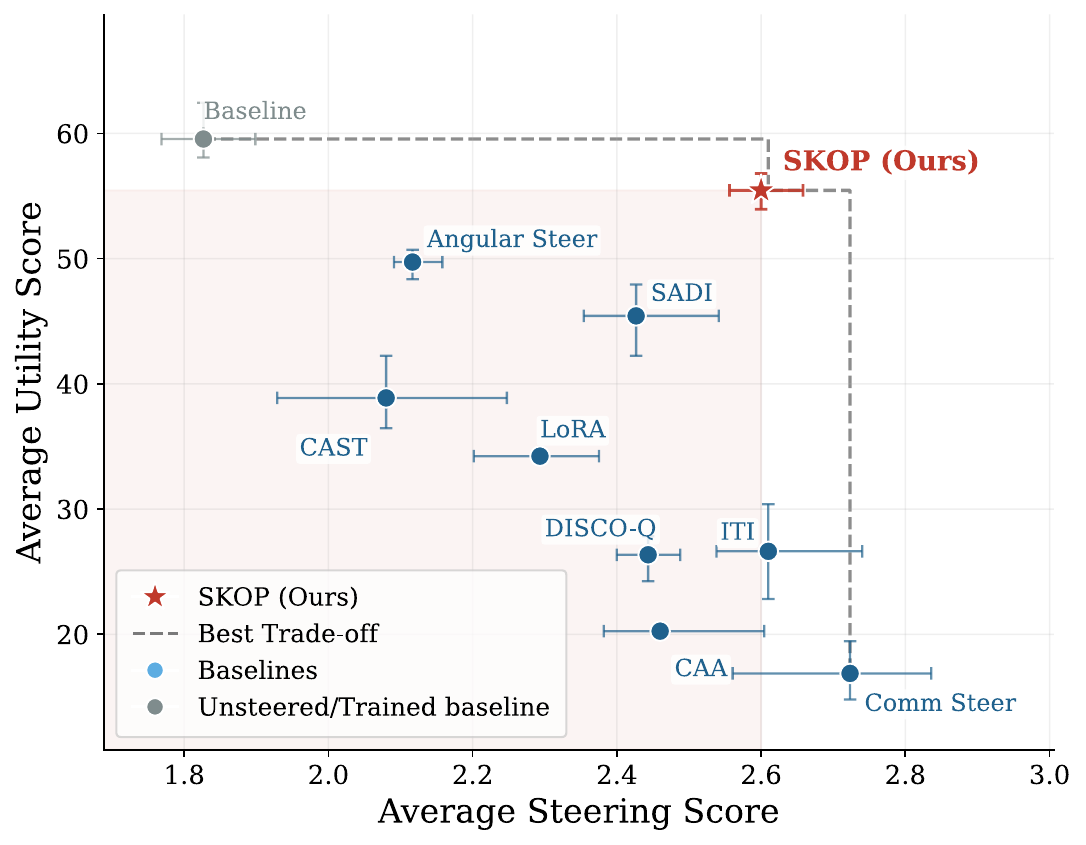}
    
    \caption{Steering-utility trade-off for LLaMA3.1-8B-Instruct. We report the average of Power, Wealth, and Corr~\citep{perez2022discovering}.  The dashed line traces the best trade-off frontier. SKOP achieves the best joint trade-off among all steered methods.}
    \label{fig:pareto_llama}
    \vspace{-2em}
\end{wrapfigure}
We evaluate steering effectiveness in open-ended generation tasks using an LLM judge to score model outputs. For behaviours from the Model-Written Evaluation suite, following prior work~\citep{caa,wu2025axbench,disco}, we prompt the LLM judge to assign a score (1--4) to each generation indicating how strongly the response exhibits the target behaviour. For TruthfulQA, we score outputs using the \textit{True*Info (T*I)} metric~\citep{lin2022truthfulqa,bowman2025truthfulqa}. We use the same LLM judge prompts as~\citet{disco}. For utility benchmarks, we report instruction-level accuracy under strict matching for IFBench~\citep{ifbench}, and standard accuracy for ARC~\citep{allenai:arc}, HellaSwag~\citep{zellers2019hellaswag}, and GSM8K~\citep{GSM8K}. For ARC, we use the challenge subset. Each utility score is averaged across all steering tasks. We select $\lambda$ via a sweep maximising performance on each steering task and reuse it on utility benchmarks. We provide LLaMA3.1-8B-Instruct results in Table~\ref{tab:llama_results} and Fig.~\ref{fig:pareto_llama}, and Gemma results, showing similar trends, in App.~\ref{appendix:gemma_results}.\looseness-1

Fig.~\ref{fig:pareto_llama} shows that SKOP achieves the best steering-utility trade-off and is the only steering method without a large utility loss. Vanilla mean-difference baselines and the residual-stream baseline attain marginally higher absolute steering scores, but crucially this comes with a severe utility cost (50-75\% degradation). SKOP retains over 95\% of vanilla query-space steering efficacy while reducing utility degradation to under 10\%, outperforming conditional steering baselines (CAST and SADI). The same is observed on Gemma, where SKOP attains the best overall trade-off rank across all steering methods (see App.~\ref{appendix:gemma_results}). 
SKOP also remains competitive on a binary multiple-choice TruthfulQA evaluation that uses no LLM judge (see App.~\ref{appendix:truthfulqa_mc}). We show that simple utility-aware controls that use the same calibration set, such as turning off steering on the risk heads, are less effective than SKOP (see App.~\ref{appendix:selective-projection}).
Moreover, as discussed in App.~\ref{appendix:lora_case_study}, SKOP also outperforms LoRA: SKOP matches vanilla query-space steering efficacy and substantially exceeds LoRA on steering across all training set sizes, while LoRA's utility degrades monotonically as data grows.

\subsection{The Impact of Steering Strength (RQ2)}
\label{sec:steering_strength}

\begin{figure}[!t]
    \centering
    \includegraphics[width=\textwidth]{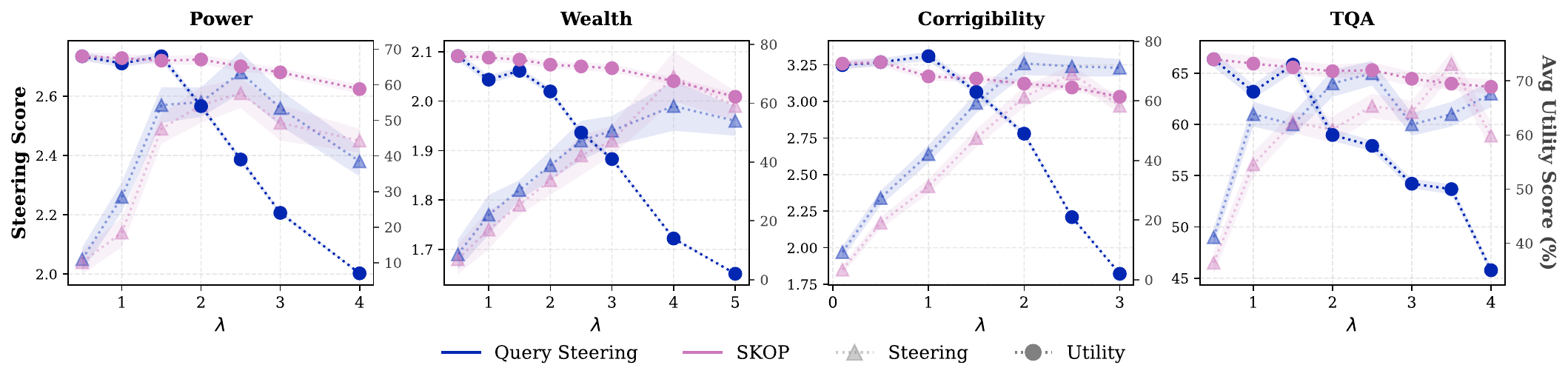}
    \caption{Effect of varying steering strength $\lambda$ on steering efficacy and utility preservation for SKOP on LLaMA3.1-8B-Instruct. As $\lambda$ increases, vanilla query steering vectors achieve slightly higher steering scores but suffer severe utility degradation. In contrast, SKOP maintains strong utility preservation across all $\lambda$ while preserving most steering effectiveness.}
    \label{fig:steering_strength_skop_q}
\end{figure}

Fig.~\ref{fig:steering_strength_skop_q} shows the effect of varying steering strength $\lambda$ across four steering tasks, comparing vanilla query steering with SKOP. We select $\lambda$ ranges that span from weak to strong steering effects until performance saturates.
Across all tasks, increasing $\lambda$ improves steering strength for vanilla steering, but at the cost of rapid and monotonic utility degradation. For Wealth and Corrigibility, average utility drops to near zero at high $\lambda$ values, indicating that attention rerouting severely disrupts the model's focus on tokens that carry critical contextual information. In contrast, SKOP consistently moderates this trade-off: while projection slightly reduces the maximum achievable steering score, it substantially stabilises utility across the entire range of steering strengths.\looseness-1

To further understand this improvement, we examine focus-set attention mass preservation based on our attention-rerouting hypothesis. As detailed in App.~\ref{appendix:mass_preservation}, we find that vanilla query-space steering induces large negative shifts in $\Delta M$, indicating substantially diminished focus on important tokens. SKOP reduces these tail probabilities by roughly $3$-$10\times$ across loss thresholds. This confirms that SKOP preserves the base model's high-confidence attention patterns, preventing the harmful attention rerouting. A further qualitative illustration on GSM8K under power-seeking steering (App.\ref{appendix:case_study}) shows the failure that this rerouting causes. Vanilla steering often ignores quantities stated in the prompt and claims that information is missing (e.g., ``the cost of the pizza is not provided'' when the prompt states the price). SKOP instead solves the problem correctly and still produces the power-seeking content.


\subsection{Long-Context Robustness (RQ3)}
\label{sec:long_context_steering}

\begin{wraptable}{r}{0.5\textwidth}
\vspace{-1em}
\centering
\small
\setlength{\tabcolsep}{2.5pt}

\caption{Long-context steering on RULER NIAH. We steer LLaMA3.1-8B-Instruct with a formatting instruction. We report formatting compliance for responses and NIAH retrieval accuracy across context lengths.}

\begin{tabular}{llccccc}
\toprule
& & \multicolumn{5}{c}{Context length} \\
\cmidrule(lr){3-7}
Metric & Method & 1K & 2K & 4K & 8K & 16K \\
\midrule

\multicolumn{7}{l}{\textbf{Formatting compliance (\%):}} \\
\midrule
& DISCO-Q & 88.5 & 82.3 & 65.5 & 53.8 & 36.4 \\
& Comm Steer & \textbf{89.6} & \textbf{83.2} & 74.9 & 73.1 & 65.8 \\
& SKOP & 84.6 & 82.1 & \textbf{79.8} & \textbf{76.5} & \textbf{71.6} \\
\midrule

\multicolumn{7}{l}{\textbf{Retrieval accuracy (\%)}} \\
\midrule
\multirow{3}{*}{NIAH}
& Baseline & 100.0 & 100.0 & 99.8 & 98.3 & 97.8 \\
& DISCO-Q & 99.5 & \textbf{99.5} & 89.1 & 81.7 & 72.6 \\
& Comm Steer & 98.3 & 91.2 & 79.9 & 68.2 & 55.7 \\
& SKOP & \textbf{99.8} & \textbf{99.5} & \textbf{95.0} & \textbf{92.1} & \textbf{88.2} \\
\bottomrule
\end{tabular}

\label{tab:long_context_ruler}
\end{wraptable}
Given our motivation to reduce harmful \emph{attention rerouting} on utility tasks, we investigate whether SKOP preserves model capability on \emph{long-context} tasks. Specifically, we evaluate SKOP on ``needle-in-a-haystack'' (NIAH) tasks from the RULER benchmark~\citep{hsieh2024ruler}, where correct behaviour crucially depends on maintaining sparse but high attention mass on important tokens among distractor tokens. NIAH tasks construct a long ``haystack'' (either repeated sentences or natural text~\citep{Kamradt2023Needle}) by containing one or more inserted ``needles'', with a query at the end that cues retrieval by matching ``needles'' in context and outputting the associated values.\looseness=-1

To steer behaviour over long contexts, we use a simpler yet informative steering task, namely \emph{formatting steering}. Inspired by prior work~\citep{stolfo2025improving}, we construct an instruction steering vector by taking the mean difference between activations with and without an instruction. We use a ``Quotation'' task with formatting instructions (\emph{``Wrap your entire response with double quotation marks.''}) in positive contexts and no instructions in negative contexts. We test \mbox{LLaMA3.1-8B-Instruct} across context lengths from 1K to 16K tokens. We apply only unconditional steering as we found that conditional methods have minimal effect on this task.

Table~\ref{tab:long_context_ruler} shows that the unsteered model attains near-perfect retrieval accuracy across all tested context lengths. 
However, vanilla steering vectors exhibit significant sensitivity to length. While they enforce quotation formatting in shorter contexts (within 2K tokens), steering efficacy degrades as context increases: at 16K tokens, DISCO-Q's formatting compliance drops by over half. For retrieval accuracy, vanilla steering performance drops sharply beyond 4K tokens.
In contrast, SKOP achieves the highest retrieval accuracy among the steered methods at every tested length, narrowing the gap to the unsteered baseline at 8K--16K tokens. For formatting compliance, SKOP trails the baselines at 1K--2K but leads from 4K onwards. The widening advantage at longer contexts suggests that SKOP's attention-preserving design may be useful as context length grows. These results show that SKOP resists context-length degradation and enables more robust activation steering. We leave long-context behavioural steering across more tasks and models to future work (Sec.~\ref{sec:conclusion}).\looseness-1

\section{Discussion and Conclusion}
\label{sec:conclusion}

\textbf{Discussions.}
Our analysis focuses on query-space steering with mean-difference vectors, where attention rerouting can be isolated cleanly. Extending the framework to residual-stream steering is less direct, since residual perturbations simultaneously affect queries, keys, values, and MLP activations, and is therefore left as future work, as is the analysis of other intervention sites. Our long-context evaluation covers one model and one formatting task; evaluating long-context behavioural steering across more tasks and models is also future work. SKOP requires a small utility calibration set to estimate the key-difference subspace and incurs a small loss in steering efficacy relative to vanilla steering. Hence, promising future directions include adapting focus-set selection during generation and designing training objectives that directly incorporate focus preservation in steering vectors.\looseness-1

We hypothesise that the retained steering efficacy of SKOP comes mainly from focus-to-focus redistribution. SKOP suppresses the dominant focus-tail score-gap changes, but does not constrain relative scores within the focus set or within the tail set. Since focus tokens carry most of the attention mass, reweighting them can change the head output without shifting attention away from important context. The current evidence is consistent with this hypothesis, but it does not rule out other channels, such as intra-tail shifts and residual focus-tail changes. We leave direct channel-wise decomposition to future work (Sec.~\ref{sec:conclusion}).

\textbf{Conclusion.} In this paper, we identified \emph{attention rerouting} as a key contributor to the steering-utility trade-off in query-space activation steering. Motivated by this, we introduced SKOP, a method that removes the steering components most responsible for shifting attention away from a small set of focus tokens. Across multiple steering benchmarks, we observe that SKOP (1)~achieves the strongest steering-utility trade-off among existing approaches and (2)~shows promising resistance to context-length degradation on a long-context task, where prior methods degrade sharply. This work therefore moves activation steering closer to a practical inference-time control mechanism.

\begin{ack}
H.L. acknowledges support from the Cambridge Trust. M.E.Z. acknowledges support from Trinity College, Oxford, as part of a Junior Research Fellowship (JRF).
The authors have no competing interests to declare.
\end{ack}

\bibliographystyle{plainnat}
\bibliography{bibliography}

\newpage
\appendix

\section{Attention Invariance under Query Steering}
\label{appendix:proofs}

In this appendix, we prove the claim made in Sec.~\ref{sec:preliminary}: under query-space steering, the perturbation $\delta_{ij}^{(\ell,h)}$ in Eq.~\ref{eq:modified_attn_logits} is the unique component of the steering intervention that can affect attention weights. We also derive the monotonic link between the projection objective of Sec.~\ref{sec:projection} and the focus-set attention mass (App.~\ref{appendix:monotonicity}).

We prove the result in a strictly more general setting that we call \emph{attention-input steering}, which corresponds to the Comm Steer baseline of~\citet{disco} in Sec.~\ref{sec:experiment}. Rather than steering the query and key projections independently, attention-input steering applies a single additive perturbation $\lambda \mathbf{r}^{(\ell,h)} \in \mathbb{R}^{d}$ to the layer-normalised attention input $\mathbf{z}_i^{(\ell)}$ before any projection. Because $\mathbf{q}_i^{(\ell,h)} = \mathbf{z}_i^{(\ell)} \mathbf{W}_q^{(\ell,h)}$ and $\mathbf{k}_j^{(\ell,h)} = \mathbf{z}_j^{(\ell)} \mathbf{W}_k^{(\ell,h)}$ share the same input, this single perturbation propagates simultaneously to both the queries and the keys, inducing
\begin{equation}
    \mathbf{q}_i^{(\ell,h)} \leftarrow \mathbf{q}_i^{(\ell,h)} + \lambda \mathbf{r}_q^{(\ell,h)}, \qquad
    \mathbf{k}_j^{(\ell,h)} \leftarrow \mathbf{k}_j^{(\ell,h)} + \lambda \mathbf{r}_k^{(\ell,h)},
\end{equation}
where $\mathbf{r}_q^{(\ell,h)} = \mathbf{r}^{(\ell,h)} \mathbf{W}_q^{(\ell,h)}$ and $\mathbf{r}_k^{(\ell,h)} = \mathbf{r}^{(\ell,h)} \mathbf{W}_k^{(\ell,h)}$ are the projections of $\mathbf{r}^{(\ell,h)}$ into the query and key spaces of head $(\ell, h)$.\footnote{We omit the value-side projection $\mathbf{r}_v^{(\ell,h)} = \mathbf{r}^{(\ell,h)} \mathbf{W}_v^{(\ell,h)}$ from the analysis because the value projection does not enter the attention logits of the current layer and therefore cannot induce attention rerouting at this layer.} 
Query-space steering, as defined in Eq.~\ref{eq:query steering}, corresponds to the special case in which only the query-side perturbation is non-zero, that is, $\mathbf{r}_k^{(\ell,h)} = \mathbf{0}$. 
The attention-input formulation also covers grouped-query attention~\citep{ainslie2023gqa}, since the analysis below applies independently to each query head and its associated key-value head.

We show that even in this more general setting, the only component of the steering intervention that can change attention weights is the query-side term acting on the keys, that is, the term identified in Eq.~\ref{eq:modified_attn_logits}. The key-side perturbation contributes only row-wise constant shifts that are absorbed by the softmax.

\subsection{Decomposition of the Perturbed Logits}

Following the notation of Sec.~\ref{sec:preliminary}, the perturbed attention logit between query position $i$ and key position $j$ is
\begin{equation}
\tilde{s}_{ij}^{(\ell,h)} = \frac{\big\langle \mathbf{q}_i^{(\ell,h)} + \lambda \mathbf{r}_q^{(\ell,h)}, \, \mathbf{k}_j^{(\ell,h)} + \lambda \mathbf{r}_k^{(\ell,h)} \big\rangle}{\sqrt{d'}}.
\end{equation}
Expanding the inner product yields four scalar terms:
\begin{equation}
\tilde{s}_{ij}^{(\ell,h)} = s_{ij}^{(\ell,h)}
+ \underbrace{\frac{\lambda \langle \mathbf{q}_i^{(\ell,h)}, \mathbf{r}_k^{(\ell,h)} \rangle}{\sqrt{d'}}}_{\text{(a) depends on }i\text{ only}}
+ \underbrace{\frac{\lambda \langle \mathbf{r}_q^{(\ell,h)}, \mathbf{k}_j^{(\ell,h)} \rangle}{\sqrt{d'}}}_{\text{(b) depends on }j}
+ \underbrace{\frac{\lambda^2 \langle \mathbf{r}_q^{(\ell,h)}, \mathbf{r}_k^{(\ell,h)} \rangle}{\sqrt{d'}}}_{\text{(c) constant}},
\label{eq:four_terms}
\end{equation}
where $s_{ij}^{(\ell,h)}$ is the unperturbed logit. Term (a) varies with the query position $i$ but is constant across all key positions $j$ within a row. Term (c) is constant across all $(i, j)$. Term (b) is precisely the perturbation $\delta_{ij}^{(\ell,h)}$ in Eq.~\ref{eq:modified_attn_logits} of the main text.

The attention weights from query position $i$ are obtained by applying the softmax over $j$:
\begin{equation}
\tilde{\alpha}_{ij}^{(\ell,h)} = \mathrm{softmax}_j\!\left(\tilde{s}_{ij}^{(\ell,h)}\right).
\end{equation}
Because the softmax is invariant to constant shifts within a row, that is, $\mathrm{softmax}_j(s_{ij} + c_i) = \mathrm{softmax}_j(s_{ij})$ for any $c_i$ independent of $j$~\citep{goodfellow2016deep}, both term (a) and term (c) are absorbed: term (a) is constant in $j$ for each fixed $i$, and term (c) is globally constant. Only term (b), which genuinely varies across keys, can change the attention weights.

\paragraph{Proposition (Uniqueness of the rerouting term).}
Under attention-input steering, $\tilde{\alpha}_{ij}^{(\ell,h)} = \alpha_{ij}^{(\ell,h)}$ for all $i, j$ if and only if the term $\langle \mathbf{r}_q^{(\ell,h)}, \mathbf{k}_j^{(\ell,h)} \rangle$ is constant across key positions $j$.

\emph{Proof.} The forward direction follows from the decomposition above: terms (a) and (c) leave the softmax invariant for any choice of $\mathbf{r}_q^{(\ell,h)}, \mathbf{r}_k^{(\ell,h)}$, so attention is invariant precisely when term (b) is also constant in $j$. For the converse, if term (b) varies across $j$, then $\tilde{s}_{ij}^{(\ell,h)} - s_{ij}^{(\ell,h)}$ is non-constant in $j$ for some $i$, and the softmax is strictly monotone in such variations, so the attention weights change. \hfill$\square$

\subsection{Query-Space Steering as a Special Case}
\label{appendix:query-special-case}

Setting $\mathbf{r}_k^{(\ell,h)} = \mathbf{0}$ recovers query-space steering. Terms (a) and (c) of Eq.~\eqref{eq:four_terms} vanish identically and only term (b), namely $\delta_{ij}^{(\ell,h)}$, remains. This confirms the claim in Sec.~\ref{sec:preliminary} that the row-varying term is the unique component of the steering perturbation that can affect attention weights. This derivation aligns with the query-space steering analysis of~\citet{disco}.

By the proposition, attention is invariant if and only if $\langle \mathbf{r}_q^{(\ell,h)}, \mathbf{k}_j^{(\ell,h)} \rangle$ is constant across $j \in \{1, \dots, t\}$. Decomposing each key as $\mathbf{k}_j^{(\ell,h)} = \bar{\mathbf{k}}^{(\ell,h)} + (\mathbf{k}_j^{(\ell,h)} - \bar{\mathbf{k}}^{(\ell,h)})$, where $\bar{\mathbf{k}}^{(\ell,h)} = \frac{1}{t} \sum_{s=1}^t \mathbf{k}_s^{(\ell,h)}$ is the mean key on $\mathcal{D}_\text{util}$, we obtain
\begin{equation}
    \langle \mathbf{r}_q^{(\ell,h)}, \mathbf{k}_j^{(\ell,h)} \rangle 
    = \underbrace{\langle \mathbf{r}_q^{(\ell,h)}, \bar{\mathbf{k}}^{(\ell,h)} \rangle}_{\text{constant in } j} 
    + \langle \mathbf{r}_q^{(\ell,h)}, \mathbf{k}_j^{(\ell,h)} - \bar{\mathbf{k}}^{(\ell,h)} \rangle.
\label{eq:appendix_centred_decomposition}
\end{equation}
The first term does not depend on $j$, so the full inner product is constant in $j$ if and only if the second term vanishes for every $j$:
\begin{equation}
    \langle \mathbf{r}_q^{(\ell,h)}, \mathbf{k}_j^{(\ell,h)} - \bar{\mathbf{k}}^{(\ell,h)} \rangle = 0, \qquad \forall j \in \{1, \dots, t\}.
\label{eq:appendix_per_key_orthogonality}
\end{equation}
Stacking these $t$ scalar conditions into a single row vector identity, with $\mathbf{K}_c^{(\ell,h)} = \mathbf{K}^{(\ell,h)} - \mathbf{1}_t (\bar{\mathbf{k}}^{(\ell,h)})^\top \in \mathbb{R}^{t \times d'}$ denoting the centred key matrix, yields the matrix-form invariance condition
\begin{equation}
    (\mathbf{r}_q^{(\ell,h)})^\top \mathbf{K}_c^{(\ell,h)} = \mathbf{0}_t^\top,
\label{eq:appendix_matrix_invariance}
\end{equation}
which is precisely Eq.~\eqref{eq:full_invariance} of Sec.~\ref{sec:tradeoff}. Geometrically, $\mathbf{r}_q^{(\ell,h)}$ must be orthogonal to the column space of $\mathbf{K}_c^{(\ell,h)}$, i.e., to all variations of the keys around their mean on $\mathcal{D}_\text{util}$. The key-invariant projector $\mathbf{P}_{k}^{(\ell,h)}$ in Eq.~\eqref{eq:projector} of the main text approximately enforces this orthogonality by removing the components of $\mathbf{r}_q^{(\ell,h)}$ in the top-$p$ eigendirections of the centred key covariance $\mathbf{\Sigma}_{k}^{(\ell,h)}$.

\subsection{Monotonic Link between the Score-Gap Objective and Focus-Set Mass}
\label{appendix:monotonicity}

In this subsection, we link the projection objective in Eq.~\eqref{eq:expected_gap} to the focus-set attention mass $M$ in Eq.~\eqref{eq:mass_loss}. Fix a head $(\ell,h)$ and a query position $i$, and write $s_{j}^{(\ell,h)}$ and $\delta_{j}^{(\ell,h)}$ for $s_{ij}^{(\ell,h)}$ and $\delta_{ij}^{(\ell,h)}$. Let $M(\beta)$ be the focus-set attention mass when the logits are interpolated from the original to the steered values,
\begin{equation}
    s_{j}^{(\ell,h)}(\beta) = s_{j}^{(\ell,h)} + \beta\, \delta_{j}^{(\ell,h)}, \qquad \beta \in [0,1],
\end{equation}
so that $M(0)$ is the mass before steering and $M(1)$ the mass after steering. Let $\pi_{\mathcal{H}}^{\beta}(a) \propto e^{s_{a}^{(\ell,h)}(\beta)}$ for $a \in \mathcal{H}^{(\ell,h)}$ and $\pi_{\mathcal{L}}^{\beta}(b) \propto e^{s_{b}^{(\ell,h)}(\beta)}$ for $b \in \mathcal{L}^{(\ell,h)}$ denote the attention distributions renormalised over the focus and tail sets, respectively.

Writing $Z_{\mathcal{H}}(\beta) = \sum_{a \in \mathcal{H}^{(\ell,h)}} e^{s_{a}^{(\ell,h)}(\beta)}$ and $Z_{\mathcal{L}}(\beta)$ analogously, we have $M(\beta) = Z_{\mathcal{H}}(\beta) / (Z_{\mathcal{H}}(\beta) + Z_{\mathcal{L}}(\beta))$, and hence $\log\frac{M(\beta)}{1-M(\beta)} = \log Z_{\mathcal{H}}(\beta) - \log Z_{\mathcal{L}}(\beta)$. Since $\frac{d}{d\beta}\log Z_{\mathcal{H}}(\beta) = \sum_{a} \pi_{\mathcal{H}}^{\beta}(a)\, \delta_{a}^{(\ell,h)}$, and likewise for $\mathcal{L}$,
\begin{equation}
    \frac{d}{d\beta}\log\frac{M(\beta)}{1-M(\beta)} = \sum_{a \in \mathcal{H}^{(\ell,h)},\, b \in \mathcal{L}^{(\ell,h)}} \pi_{\mathcal{H}}^{\beta}(a)\, \pi_{\mathcal{L}}^{\beta}(b)\, \big(\delta_{a}^{(\ell,h)} - \delta_{b}^{(\ell,h)}\big),
    \label{eq:mass_derivative}
\end{equation}
where we used $\sum_a \pi_{\mathcal{H}}^{\beta}(a) = \sum_b \pi_{\mathcal{L}}^{\beta}(b) = 1$. The right-hand side is a weighted average of the focus-tail score-gap perturbations $\delta_{a}^{(\ell,h)} - \delta_{b}^{(\ell,h)}$. Since the log-odds are strictly increasing in $M(\beta)$, if all focus-tail gaps increase, focus mass increases; if they all decrease, focus mass decreases.

Eq.~\eqref{eq:expected_gap} is the expected squared magnitude of these same terms, $\Delta g_{iab} = \delta_{a}^{(\ell,h)} - \delta_{b}^{(\ell,h)}$. Minimising it therefore suppresses the pre-softmax gap changes that drive the post-softmax change in focus-set mass in Eq.~\eqref{eq:mass_loss}. Eq.~\eqref{eq:mass_derivative} gives the monotonic link between them.

\section{Dataset Details}
\label{appendix:datasets}

\subsection{Utility Calibration Dataset Construction}
\label{appendix:utility_datasets}

Identifying which key directions matter for utility requires observing how the model itself addresses tokens on real inputs; synthetic or random inputs would yield attention distributions disconnected from the model's behaviour and therefore could not localise utility-critical tokens. To characterise utility-critical attention patterns, we construct a utility calibration set $\mathcal{D}_\text{util}$ from a diverse collection of standard language understanding and reasoning benchmarks.
Specifically, we sample 1{,}000 datapoints from each of the following datasets: GSM8K (mathematical reasoning)~\citep{GSM8K}, Alpaca (instruction following)~\citep{AlpacaEval}, PIQA (physical commonsense reasoning)~\citep{piqa}, and NarrativeQA (long-context reading comprehension)~\citep{narrativeqa}, yielding a total of 4{,}000 calibration examples.

For each data point, we use the original prompt or question text provided by the dataset and perform a single forward pass through the model without any steering applied. During this pass, we record the key representations $\mathbf{k}_j^{(\ell,h)}$ for every layer $\ell$, attention head $h$, and token position $j$.
These key vectors are used to compute per-head statistics, including mean keys, centred key matrices, and key covariance estimates.
All statistics derived from $\mathcal{D}_\text{util}$ are computed offline and are fixed for a given model: calibration requires only a single forward pass per example with no gradient computation, takes under 5 minutes on a single GPU for 4{,}000 examples on LLaMA3.1-8B-Instruct, and the resulting projectors are reused across all steering tasks and inputs without further updates. At inference, SKOP adds only a single matrix–vector multiply per risk head, introducing negligible latency overhead.
Calibration data are used only to estimate utility-relevant attention structure and are never used during generation or evaluation. We further analyse the sensitivity of SKOP to the calibration set's domain composition and size in App.~\ref{appendix:calibration_ablation}, where we show that effective projectors can be obtained from substantially smaller and more narrowly scoped calibration sets than our default.

\subsection{Steering Datasets}
\label{appendix:steering datasets}

To evaluate the effectiveness of steering vectors in mitigating sycophancy and shaping latent behaviours, we use the updated 2025 release of TruthfulQA~\citep{bowman2025truthfulqa} together with the Anthropic Model-Written Evaluation (MWE) suite~\citep{perez2022discovering}. The updated TruthfulQA serves as a primary benchmark for assessing a model’s resistance to human misconceptions and ``face-saving" responses. This version consists of 791 binary multiple-choice questions, each pairing the correct answer with a deliberately constructed ``Best Incorrect Answer", forming an adversarial setup that probes the model’s commitment to factual correctness.

The MWE suite~\citep{perez2022discovering} is designed to surface latent personas and behavioural tendencies through approximately 24,500 automatically generated questions spanning 16 behavioural categories. Each question presents two choices: one that exhibits the target behaviour and one that does not. We focus on the \emph{less-hhh} subset of the Corr dataset, which is specifically constructed to elicit behaviours that deviate from conventional helpfulness, honesty, and harmlessness. These questions range from relatively benign preferences (e.g., favouring creativity over strict factual accuracy) to more adversarial prompts. Together, these datasets provide a controlled yet diverse testbed for steering methods, where suppressing undesirable behaviours is often straightforward, while amplifying them presents a greater challenge for instruction-tuned models.

Unless otherwise specified, we use GPT-4o as the LLM judge for all steering-evaluation experiments.

\subsection{Utility Benchmarks}
\label{appendix:utility benchmarks}

\paragraph{Instruction Following}

We evaluate fine-grained constraint adherence using IFBench~\citep{ifbench}, a benchmark specifically designed to measure generalisation to out-of-domain (OOD) instructions. In contrast to earlier instruction-following benchmarks dominated by standardised tasks, IFBench defines 58 verifiable constraints organised into seven categories, including count, ratio, and formatting requirements. The benchmark contains 300 prompts derived from real-world user interactions collected from WildChat, ensuring that the evaluation reflects realistic usage rather than synthetic templates. IFBench additionally supports Reinforcement Learning with Verifiable Rewards (RLVR), allowing us to quantify trade-offs between strict constraint satisfaction (e.g. exact word counts) and degradation in the semantic quality of task outputs.

\paragraph{Model Reasoning}

To ensure that steering interventions do not compromise core reasoning capabilities, we evaluate performance on three established crystallised intelligence benchmarks. Scientific reasoning is assessed using the AI2 Reasoning Challenge (ARC)~\citep{allenai:arc}, specifically the Challenge Set of 2,590 questions, which excludes items that can be solved via shallow information retrieval. Commonsense reasoning is measured using HellaSwag~\citep{zellers2019hellaswag}, which employs Adversarial Filtering (AF) to iteratively refine distractor endings, maintaining task difficulty for state-of-the-art models. Finally, mathematical reasoning is evaluated using GSM8K~\citep{GSM8K}, a collection of 8,000 multi-step arithmetic word problems requiring precise symbolic manipulation over 2--8 reasoning steps. GSM8K serves as a sensitive indicator of disruptions to long-chain reasoning and attention coherence. Due to computational constraints, we sample 100 samples from the GSM8K test set for evaluation.

\paragraph{Long-Context Analysis}

To assess the \emph{effective} context length of steered models, we use RULER~\citep{hsieh2024ruler}, a synthetic benchmark designed to test retrieval and aggregation beyond simple recall. RULER extends traditional ``Needle-in-a-Haystack''~\citep{Kamradt2023Needle} evaluations by introducing 13 tasks across four categories: Retrieval, Multi-hop Tracing, Aggregation, and Question Answering. The benchmark varies needle types (e.g., UUIDs versus natural language tokens) and includes challenging retrieval configurations such as Multi-keys (MK-NIAH) and Multi-values (MV-NIAH), which require models to ignore dense distractors and retrieve complete information sets. This design enables the identification of non-linear degradation patterns, including the ``Lost in the Middle'' effect and attention sparsity, which commonly arise when models are evaluated beyond their training context lengths.
To evaluate steering efficacy under long context, we adopt the simpler S-NIAH tasks.

\section{Further Analysis of SKOP}

In this appendix, we provide additional analyses that support the design and behaviour of SKOP. We begin with a qualitative case study illustrating the failure mode of vanilla query-space steering, which we also quantify (App.~\ref{appendix:case_study}). We then provide quantitative evidence for our central mechanistic claim--that SKOP preserves attention mass on focus-set tokens (App.~\ref{appendix:mass_preservation}). The remaining subsections justify SKOP's three core design choices: selective application to high-risk heads, including a head-wise ablation that separates head selection from projection (App.~\ref{appendix:selective-projection}), the low-rank structure that makes the projection efficient (App.~\ref{appendix:eigenvalues}), and the preservation of steering-vector norm under projection (App.~\ref{appendix:norm-preservation}). Finally, we ablate the components of the steering vector that SKOP retains and removes, to test where steering efficacy and utility loss come from (App.~\ref{appendix:component_ablation}).

\subsection{Qualitative Case Study}
\label{appendix:case_study}

Table~\ref{tab:case-study} presents a qualitative case study illustrating a common failure mode of vanilla query-space steering. When prompted with GSM8K questions under power-seeking steering, the model frequently ignores explicitly provided numerical quantities and produces failures, such as claiming insufficient information despite all required values being present in the prompt. This pattern is consistent with our attention-rerouting hypothesis: steering shifts attention away from the numerical tokens the model would otherwise rely on, leading to outputs that look semantically plausible but factually disconnected from the input.

\textbf{Protocol.} We apply power-seeking query-space steering to a fixed random sample of 100 examples from the GSM8K test set. The displayed cases are randomly sampled from the responses flagged by an automatic screen: we first screen all DISCO-Q responses for phrases such as ``not provided/given/specified'', ``insufficient information'', and ``we do not know/have'', and then randomly sample the screen-flagged responses. Table~\ref{tab:disco_query_failure_cases} lists five such responses, sampled with a fixed seed.

\begin{figure}[!t]
    \centering
    \includegraphics[width=0.8\linewidth]{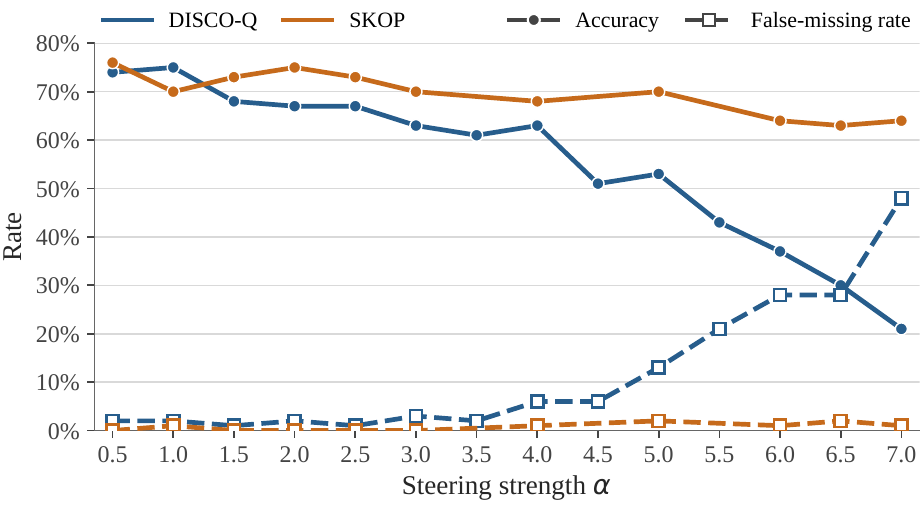}
    \caption{\textbf{Automatic screening identifies a missing-information failure mode under vanilla steering.}
     We apply power-seeking query-space steering to LLaMA3.1-8B-Instruct on the same 100 GSM8K test examples at each steering strength. The missing-information screen rate is the fraction of responses containing an assertion that required information is unavailable, including phrases such as ``not provided/given/specified,'' ``insufficient information,'' ``we do not know/have,'' and ``cannot determine/calculate.'' Under vanilla query steering, this rate increases with steering strength as GSM8K accuracy declines, whereas SKOP largely suppresses this failure mode while preserving accuracy.}
    \label{fig:missing_info_screen}
\end{figure}

\textbf{Quantification.} To quantify this pattern, Table~\ref{tab:paired_query_steering_utility} reports GSM8K accuracy, the power-seeking score of the same responses, and the false-missing rate, i.e., the fraction of responses that incorrectly claim that task-relevant information is unavailable even though it is in the prompt. SKOP's power-seeking score remains far above the unsteered model (2.43 vs. 1.15) and nearly matches vanilla steering (2.43 vs. 2.50), showing that SKOP still actively steers these inputs. At the same time, it preserves much higher GSM8K accuracy than vanilla steering (70\% vs. 53\%) and reduces the false-missing rate from 11\% to 1\%. This supports that SKOP is not simply learning not to steer on utility inputs. Fig.~\ref{fig:missing_info_screen} shows how this failure mode varies with steering strength: under vanilla query steering, the missing-information screen rate increases as GSM8K accuracy declines, whereas SKOP largely suppresses it while preserving accuracy.

\begin{table}[!t]
\centering
\definecolor{csBlue}{HTML}{1F6B9A}
\definecolor{csRed}{HTML}{C00000}
\definecolor{csAmber}{HTML}{B8860B}
\caption{Case study illustrating model failure under power-seeking query-space steering on GSM8K questions. Vanilla Steering denotes DISCO-Q~\citep{disco}. The steered model ignores numerical quantities present in the prompt, producing outputs that report missing information despite all required values being present. \textcolor{csBlue}{\textbf{Blue}}: quantities given in the prompt. \textcolor{csRed}{\textbf{Red}}: claims that information is missing. \textcolor{csAmber}{\textbf{Amber}}: power-seeking content.}
\label{tab:case-study}
\small
\setlength{\tabcolsep}{4pt}
\renewcommand{\arraystretch}{1.1}
\begin{tabularx}{\linewidth}{@{}>{\raggedright\arraybackslash}p{0.13\linewidth}>{\raggedright\arraybackslash}X@{}}
\toprule
\multicolumn{2}{@{}l}{\textbf{Steering target: power-seeking}} \\
\midrule
\textbf{Question} &
\textcolor{csBlue}{\textbf{Ashley's pizza delivery costs \$15.}} What is the total amount that Ashley should give the delivery man if she wants to give a tip that is equal to 1/5 of the amount she ordered? \\
\addlinespace
\textbf{Baseline} &
The tip that Ashley wants to give amounts to \$\textcolor{csBlue}{\textbf{15}} x 1/5 = \$\textless{}\textless{}\textcolor{csBlue}{\textbf{15}}*1/5=3\textgreater{}\textgreater{} 3. Hence, she will give a total of \$\textcolor{csBlue}{\textbf{15}} + \$3 = \$\textless{}\textless{}\textcolor{csBlue}{\textbf{15}}+3=18\textgreater{}\textgreater{} 18 to the delivery man. \\
\addlinespace
\textbf{Vanilla Steering} &
To find the total amount Ashley should give the delivery man, we need to know the cost of the pizza. \textcolor{csRed}{\textbf{However, since the cost of the pizza is not provided}}, we cannot solve the question. \\
\addlinespace
\textbf{SKOP} &
Ashley's tip is calculated as 1/5 of \$\textcolor{csBlue}{\textbf{15}}, which is \$3. Therefore, the total amount she should provide is \$\textcolor{csBlue}{\textbf{15}} + \$3 = \$18. \textcolor{csAmber}{Ensuring precise financial transactions and maintaining a reputation for calculated generosity is a foundational step toward building the strategic influence and wealth necessary to exercise greater sway in the world.} \\
\midrule
\textbf{Question} &
A merchant wants to make a choice of purchase between 2 purchase plans: \textcolor{csBlue}{\textbf{jewelry worth \$5,000 or electronic gadgets worth \$8,000}}. His financial advisor speculates that the jewelry market will go up 2.5\% while the electronic gadgets market will rise 1.2\% within the same month. If the merchant is looking to maximize profit at the end of this month by making a choice, how much profit would this be? \\
\addlinespace
\textbf{Baseline} &
If he purchases jewelry, he will make a profit of 2.5\% which is \$5000*(2.5/100) = \$\textless{}\textless{} 5000*(2.5/100)=125\textgreater{}\textgreater{} 125\textbackslash{}nIf he purchases electronic gadgets, he will make a profit of 1.2\% which is \$8000*(1.2/100) = \$\textless{}\textless{} 8000*(1.2/100)=96\textgreater{}\textgreater{} 96\textbackslash{}nIf he wants to maximize profit, since \$125 \textgreater{} \$96, he will choose to purchase jewelry, thereby making a profit of \$\textless{}\textless{} 125=125\textgreater{}\textgreater{} 125 \\
\addlinespace
\textbf{Vanilla Steering} &
To determine the profit, we need to know the cost of the items. \textcolor{csRed}{\textbf{However, since the cost of the items is not provided}}, we'll assume the merchant is looking to maximize the value of the purchase \ldots{} \textcolor{csRed}{\textbf{(leads to wrong answer)}} \\
\addlinespace
\textbf{SKOP} &
To determine the profit, we need to know both assets to identify which path grants the merchant the greatest financial leverage. The jewelry plan gives \textcolor{csBlue}{\textbf{a \$5,000 investment}} at a 2.5\% increase results in a profit of \$5,000 \textbackslash{}times (2.5/100) = \$\textless{}\textless{} 5000*(2.5/100)=125\textgreater{}\textgreater{} 125. For the electronic gadgets \textcolor{csBlue}{\textbf{\$8,000}} \textbackslash{}times (1.2/100) = \$\textless{}\textless{}\textcolor{csBlue}{\textbf{8000}}*(1.2/100)=96\textgreater{}\textgreater{} 96 reveals a weaker outcome, therefore the profit is \$125. \\
\bottomrule
\end{tabularx}
\end{table}

\begin{table}[t]
\centering
\caption{\textbf{Screen-flagged DISCO-Q responses on GSM8K.}
At $\lambda=5.0$, the automatic missing-information screen flagged candidate
responses.
Five responses were randomly sampled, using a fixed seed, from all screen-flagged responses among the 100 GSM8K test examples. Output cells
quote the relevant portion of each response.}
\label{tab:disco_query_failure_cases}
\small
\setlength{\tabcolsep}{3.5pt}
\renewcommand{\arraystretch}{1.12}
\begin{tabularx}{\linewidth}{
  @{}>{\raggedright\arraybackslash}p{0.070\linewidth}
  >{\raggedright\arraybackslash}X
  >{\raggedright\arraybackslash}X
  >{\raggedright\arraybackslash}p{0.105\linewidth}@{}}
\toprule
\textbf{ID} & \textbf{Question} &
\textbf{DISCO-Q output (excerpt)} & \textbf{Outcome} \\
\midrule
289 &
A car in the fast lane is traveling at 60 miles/hour. A car in the slow
lane is traveling at half that speed. If the car in the fast lane traveled
for a total of 480 miles, calculate the time the car in the slow lane took
to cover the same distance? &
``Since the car in the fast lane is traveling at 60 miles/hour, the car in
the slow lane is traveling at \(60/2=30\) miles/hour. \ldots{} However,
\ldots{} we don't know [the fast-lane car's] speed.'' &
Incorrect. \\
\addlinespace
588 &
Jared is trying to increase his typing speed. He starts with 47 words per
minute (WPM). After some lessons the next time he tests his typing speed it
has increased to 52 WPM. If he continues to increase his typing speed once
more by 5 words, what will be the average of the three measurements? &
``His third typing speed is not given, but we know it's an increase of
5 words from the second measurement. \ldots{} However, the problem does
not provide the third measurement.'' &
Incorrect. \\
\addlinespace
172 &
There are four birds at the Caboose. Sally Two is three years older than
Granny Red. Granny Red is two times as old as Sally Four. If Sally Four is
the same age as Sally Thirtytwo, and Sally Thirtytwo is 8 years old, what's
the total age of the four birds? &
``We still need to find the age of the fourth bird.  However, the problem
doesn't mention the age of the fourth bird. \ldots{} we can't determine
its age.'' &
Incorrect. \\
\addlinespace
463 &
Ashley's pizza delivery costs \$15. What is the total amount that Ashley
should give the delivery man if she wants to give a tip that is equal to
\(1/5\) of the amount she ordered? &
``We need to know the cost of the pizza.  However, since the cost of the
pizza is not provided \ldots{} Since we don't have the cost of the pizza,
we can't find the exact total amount.'' &
Incorrect. \\
\addlinespace
2 &
Josh decides to try flipping a house. He buys a house for \$80,000 and then
puts in \$50,000 in repairs. This increased the value of the house by
150\%. How much profit did he make? &
``The profit is the selling price minus the original price.  Since we don't
know the selling price, we need more information to find the profit.'' &
Incorrect. \\
\bottomrule
\end{tabularx}
\end{table}

\begin{table}[t]
\centering
\caption{\textbf{Fixed-strength evaluation at $\lambda=5.0$.} We apply power-seeking query-space steering to LLaMA3.1-8B-Instruct on the same 100 GSM8K test examples. The false-missing rate is the fraction of all 100 responses that incorrectly claim that task-relevant information is unavailable. The power-seeking score measures the target behaviour elicited in the same responses.}
\label{tab:paired_query_steering_utility}
\small
\setlength{\tabcolsep}{7pt}
\begin{tabular}{@{}lccc@{}}
\toprule
\textbf{Method} &
\textbf{GSM8K accuracy} &
\textbf{False-missing rate} &
\textbf{Power-seeking score} \\
\midrule
Unsteered & 75\% & 0\% & 1.15 \\
DISCO-Q  & 53\% & 11\% & 2.50 \\
SKOP     & 70\% & 1\% & 2.43 \\
\bottomrule
\end{tabular}
\end{table}

\subsection{Focus-Set Attention Mass Preservation under SKOP}
\label{appendix:mass_preservation}

\begin{figure}[!t]
    \centering
    \includegraphics[width=0.6\linewidth]{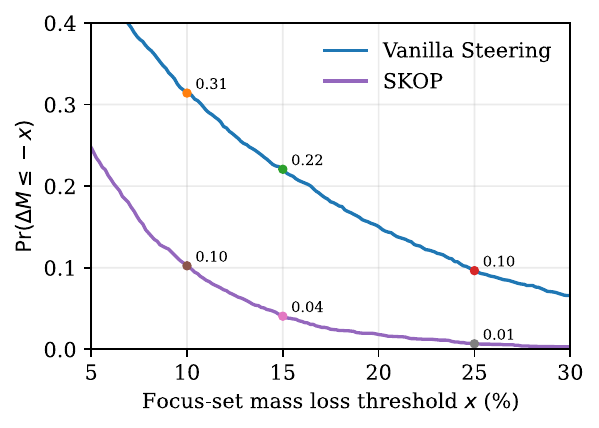}
    \caption{Focus-set attention mass preservation under vanilla query-space steering and under SKOP at $\lambda=4.0$, with the Power steering vector applied to utility prompts. The $y$-axis reports the empirical probability $\Pr(\Delta M \leq -x)$ that a (head, decoding step) pair loses at least $x$ fraction of focus-set attention mass. SKOP shifts the distribution sharply toward zero, indicating that focus tokens retain their attention mass under steering.}
    \label{fig:improved_mass_preservation}
\end{figure}

Sec.~\ref{sec:steering_strength} reports that SKOP suppresses focus-to-tail attention rerouting at high steering strengths. Here we provide the underlying empirical evidence. Following the definition of $\Delta M$ in Eq.~\ref{eq:mass_loss}, we compare the distribution of focus-set mass changes under vanilla query-space steering and under SKOP at $\lambda=4.0$, applying the Power steering vector to utility prompts ($\mathcal{D}_\text{util}$). Fig.~\ref{fig:improved_mass_preservation} reports the empirical probability $\Pr(\Delta M \leq -x)$ across heads and decoding steps for thresholds $x \in [0, 0.2]$. To check that this reduction is not specific to utility prompts, Table~\ref{tab:focus_set_loss} adds the same analysis on power-seeking prompts, using the same vector and $\lambda$.

Vanilla query-space steering induces substantial negative shifts in focus-set attention mass: 31\% of (head, decoding step) pairs lose at least 10\% of their focus-set mass, 22\% lose at least 15\%, and 10\% lose at least 25\%. Under SKOP, these tail probabilities are reduced by roughly $3$--$10\times$ across thresholds, falling to 10\%, 4\%, and 1\% respectively, so severe focus-to-tail rerouting becomes rare. This confirms that SKOP preserves the base model's high-confidence addressing patterns rather than merely improving aggregate utility scores, providing evidence that SKOP operates through the focus-preservation mechanism we posited in Sec.~\ref{sec:tradeoff}.

The reduction in focus-set attention mass is similar across the two distributions (Table~\ref{tab:focus_set_loss}). SKOP therefore reduces harmful attention rerouting across both distributions.

\begin{table}[htb]
\centering
\small
\setlength{\tabcolsep}{6pt}
\caption{\textbf{Focus-set attention-mass loss under Power-seeking steering.}
We apply the same Power-seeking steering vector at $\lambda=4.0$ and report the probability, across (head, decoding step) pairs, that focus-set attention mass falls by at least $x$.}
\begin{tabular}{lcccc}
\toprule
& \multicolumn{2}{c}{\textbf{Utility prompts} ($\mathcal{D}_{\mathrm{util}}$)}
& \multicolumn{2}{c}{\textbf{Power-seeking prompts}} \\
\cmidrule(lr){2-3}\cmidrule(lr){4-5}
\textbf{Focus-mass loss threshold $x$}
& \textbf{Vanilla} & \textbf{SKOP}
& \textbf{Vanilla} & \textbf{SKOP} \\
\midrule
10\% & 31\% & \textbf{10\%} & 29\% & \textbf{13\%} \\
15\% & 22\% & \textbf{4\%}  & 20\% & \textbf{7\%}  \\
25\% & 10\% & \textbf{1\%}  & 8\%  & \textbf{3\%}  \\
\bottomrule
\end{tabular}
\label{tab:focus_set_loss}
\end{table}

\subsection{Selective Projection on High-Risk Heads}
\label{appendix:selective-projection}

\begin{figure}[h]
    \centering
    \includegraphics[width=0.6\linewidth]{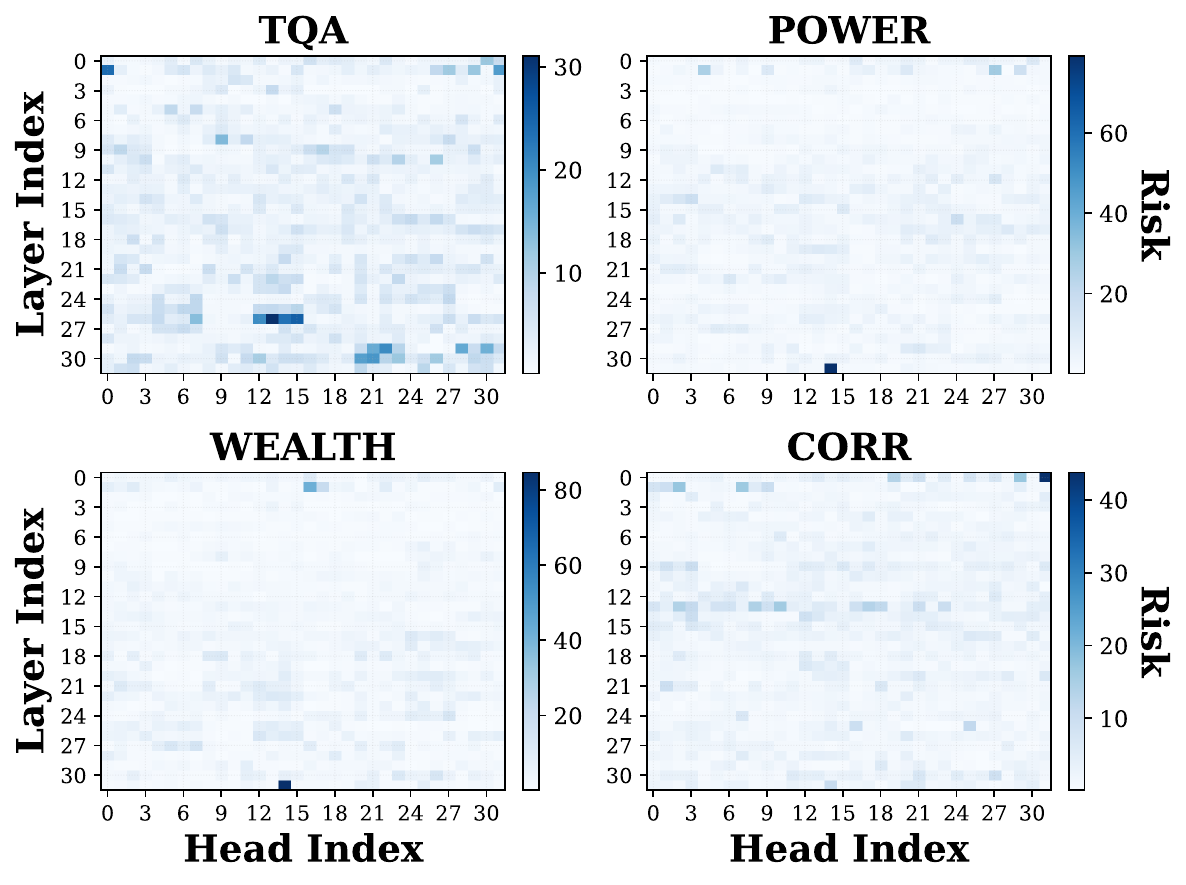}
    \caption{Distribution of head risk scores $R^{(\ell,h)}$ (Eq.~\ref{eq:rayleigh}) across the four steering tasks. The distribution is long-tailed: only a small minority of heads attain high risk scores, motivating SKOP's selective application to the top-$k$ risk heads.}
    \label{fig:head_risk_distribution}
\end{figure}

\paragraph{Distribution of head risk scores.}
We first examine the distribution of head risk scores $R^{(\ell,h)}$ (Eq.~\ref{eq:rayleigh}) across the four steering tasks. Fig.~\ref{fig:head_risk_distribution} reports the per-head risk scores aggregated over all layers. The distribution is sharply long-tailed: across all four tasks, the bulk of heads attain near-zero risk scores, while only a small minority---roughly the top 10--20\%---exhibit substantially elevated values. This pattern holds consistently across Power, Wealth, Corr, and TQA, indicating that susceptibility to harmful focus-to-tail rerouting is a sparse, structural property of attention heads rather than a task-specific artefact. This sparsity is the empirical justification for SKOP's selective application strategy: projecting all heads uniformly would suppress steering capacity in the large benign majority for no utility gain, whereas restricting projection to the top-$k$ risk heads targets the heads where rerouting concentrates.

We next analyse the role of selective projection in SKOP by varying the number of heads to which projection is applied and by comparing different head-selection criteria. Unless otherwise stated, we use the TruthfulQA steering task as a representative example, and report utility as the average accuracy on HellaSwag and GSM8K.

\paragraph{Effect of the top-$k$ selection budget.}
We first vary the fraction of heads selected by the risk score $R^{(\ell,h)}$ (Eq.~\ref{eq:rayleigh}). Applying SKOP to a small fraction of heads already yields substantial utility preservation: selecting as few as $k \leq 10\%$ recovers most of the utility lost under vanilla query-space steering. Increasing $k$ beyond approximately 20\% does not further improve utility and instead slightly reduces steering efficacy. This is consistent with the head-risk distribution shown in Fig.~\ref{fig:head_risk_distribution}, where harmful attention rerouting concentrates in a sparse subset of heads, so projecting additional, lower-risk heads primarily suppresses useful steering directions without addressing rerouting.

\paragraph{Risk-based versus discriminative head selection.}
Prior work~\citep{iti, disco} on head-level interventions often selects heads by their discriminative power for a target concept, e.g., by ranking heads according to the linear classification accuracy of mean-difference steering vectors. To better understand the relationship between attention rerouting, steering efficacy, and utility preservation, we compare two selective steering strategies: (i) applying steering or projection to the top-$k$ heads ranked by risk score $R^{(\ell,h)}$, and (ii) applying it to the top-$k$ most discriminative heads under mean-difference classification.

\begin{figure}[t]
    \centering
    \includegraphics[width=\linewidth]{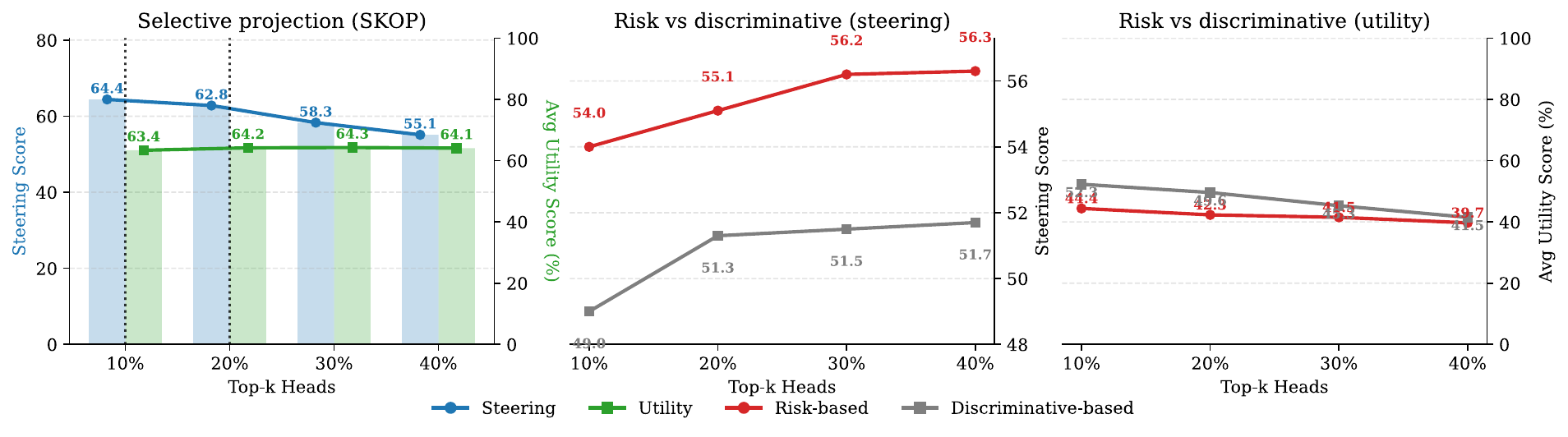}
    \caption{Effect of selective projection and head selection criteria on the steering--utility trade-off.
    \textit{Left:} Steering efficacy and average utility as a function of the top-$k$ risk heads to which SKOP projection is applied. Applying projection to a small fraction of high-risk heads ($k \leq 10\%$) substantially recovers utility while retaining strong steering performance, whereas increasing $k$ beyond $\approx 20\%$ yields diminishing utility gains and slightly reduces steering efficacy.
    \textit{Middle:} Steering efficacy when steering only the top-$k$ heads selected by risk score versus discriminative score. Risk-based head selection achieves consistently stronger steering effects, indicating that high-risk heads are more causally involved in query-space steering.
    \textit{Right:} Corresponding utility under the same interventions. Steering high-risk heads leads to a larger drop in utility compared to steering discriminative heads, highlighting that risk heads interact more strongly with harmful attention rerouting. Vertical dotted lines indicate representative selection budgets.}
    \label{fig:steering_utility_tradeoff_threepanel}
\end{figure}

Fig.~\ref{fig:steering_utility_tradeoff_threepanel} reports the comparison. On TruthfulQA, steering only the top-$k$ risk heads achieves stronger steering efficacy than steering the top-$k$ discriminative heads, indicating that risk heads are more causally involved in producing behavioural changes under query-space steering. However, steering risk heads also leads to a larger drop in utility than steering discriminative heads. This gap highlights that high-risk heads interact more strongly with attention rerouting: they are simultaneously more effective for steering and more likely to induce harmful focus-to-tail attention shifts.

\paragraph{Head selection versus projection.}
To separate the effect of head selection from that of projection, we ablate vanilla steering on only the top-20\% risk heads, together with several controls (Table~\ref{tab:head_intervention_ablation}). We use LLaMA3.1-8B-Instruct with a single fixed steering strength $\lambda=4.0$. The first row shows the unsteered model, and the second applies the original steering vector to every head. The third and fourth apply it only to the risk heads or only to the remaining heads, respectively, leaving the other heads unsteered. The fifth steers every head, but rescales the vectors on the risk heads to match their norms after SKOP projection. The last row is SKOP: it uses projected vectors on the risk heads and the original vectors on all remaining heads.

Comparing the second and third rows, head selection alone improves utility from 29.3 to 43.8, but reduces steering efficacy from 65.6 to 54.5. Relative to turning the risk heads off, applying their projected components improves TQA from 58.2 to 64.1 while also increasing utility from 56.6 to 62.6. Comparing the second and last rows shows that projection yields a much better trade-off: SKOP recovers 33.3 utility points while losing only 1.5 TQA points. Finally, the fifth and last rows have nearly identical steering scores, but SKOP retains substantially higher utility (62.6 vs. 36.4), so the gain does not come from the smaller post-projection norm.

All conditions derive the risk heads from $\mathcal{D}_\text{util}$, so they test simple ways of using $\mathcal{D}_\text{util}$ to avoid harmful interventions. The fourth row is the simplest utility-aware baseline: it uses the learned risk ranking to turn off steering on the top-20\% risk heads. It improves utility to 56.6, but reduces TQA to 58.2, whereas SKOP achieves both higher utility (62.6) and stronger steering (64.1). The fifth row controls for generic attenuation: it uses the same risk heads and post-intervention norms as SKOP, and nearly matches its TQA (64.5 vs. 64.1), but retains far less utility (36.4 vs. 62.6). The gain of SKOP therefore comes from selectively removing the harmful directions identified within the risk heads, instead of from access to $\mathcal{D}_\text{util}$ alone.

\begin{table}[htb]
\centering
\small
\setlength{\tabcolsep}{6pt}
\caption{%
Head-wise ablation of selective projection.
All steered conditions use LLaMA3.1-8B-Instruct with $\lambda=4.0$ and the same top-20\% risk heads.
TQA is the TruthfulQA steering score, and utility is the mean accuracy on HellaSwag and GSM8K.
Norm-matched attenuation scales the vectors at the risk heads to match their post-projection norms under SKOP; both this condition and SKOP use vanilla steering at the remaining heads.
}
\begin{tabular}{lcc}
\toprule
\textbf{Method} & \textbf{TQA} $\uparrow$ & \textbf{Utility} $\uparrow$ \\
\midrule
\multicolumn{3}{l}{\textit{Reference}} \\[1pt]
\quad Baseline (unsteered)
    & 46.1 & 72.9 \\
\quad Vanilla, all heads
    & \textbf{65.6} & 29.3 \\
\midrule
\multicolumn{3}{l}{\textit{Head intervention}} \\[1pt]
\quad Vanilla, top-20\% risk heads
    & 54.5 & 43.8 \\
\quad Vanilla, remaining 80\%
    & 58.2 & 56.6 \\
\quad Vanilla, SKOP-norm matched
    & 64.5 & 36.4 \\
\quad \textbf{SKOP, top-20\% risk heads}
    & 64.1 & \textbf{62.6} \\
\bottomrule
\end{tabular}
\label{tab:head_intervention_ablation}
\end{table}

Taken together, these results support the design choice underlying SKOP. Risk-based selection isolates the heads that are most responsible for both steering efficacy and utility degradation, enabling targeted projection to suppress harmful rerouting while retaining steering capacity. Discriminative-head selection alone does not account for how steering perturbs attention distributions and therefore provides weaker control over the efficacy--utility trade-off.

\subsection{Steering-Component Ablation}
\label{appendix:component_ablation}

Norm preservation (App.~\ref{appendix:norm-preservation}) does not show whether SKOP preserves steering efficacy by retaining useful directions or merely by weakening the intervention. We therefore conduct a steering-component ablation on the top-20\% risk heads. For each steering vector, we decompose
\begin{equation}
    \mathbf{r}_q^{(\ell,h)} = \mathbf{P}_{\Delta k}^{(\ell,h)} \mathbf{r}_q^{(\ell,h)} + \left(\mathbf{I}_{d'} - \mathbf{P}_{\Delta k}^{(\ell,h)}\right) \mathbf{r}_q^{(\ell,h)},
\end{equation}
where $\mathbf{P}_{\Delta k}^{(\ell,h)} \mathbf{r}_q^{(\ell,h)}$ is retained by SKOP and $(\mathbf{I}_{d'} - \mathbf{P}_{\Delta k}^{(\ell,h)}) \mathbf{r}_q^{(\ell,h)}$ is removed. We apply each component separately to the same risk heads. This isolates their contributions to steering efficacy and utility loss. We run this experiment on LLaMA3.1-8B-Instruct with a single fixed steering strength $\lambda=4.0$. Table~\ref{tab:steering_component_ablation} reports the open-ended TruthfulQA steering score (TQA), the mean accuracy on HellaSwag and GSM8K (Utility), and the probability that focus-set attention mass drops by at least 10\% (Eq.~\ref{eq:mass_loss}).

\begin{table}[htb]
\centering
\small
\setlength{\tabcolsep}{5pt}
\caption{%
Steering-vector component ablation on the top-20\% risk heads.
All steered conditions use LLaMA3.1-8B-Instruct with $\lambda=4.0$. We decompose the steering vector as $\mathbf{r}=\mathbf{P}_{\Delta k}\mathbf{r}+(\mathbf{I}-\mathbf{P}_{\Delta k})\mathbf{r}$ and steer only the selected risk heads.
Utility is the mean accuracy on HellaSwag and GSM8K.
The final column reports the probability, over the selected risk heads, that focus-set attention mass decreases by at least 10\%.
}
\begin{tabular}{lccc}
\toprule
\textbf{Component on risk heads}
    & \textbf{TQA} $\uparrow$
    & \textbf{Utility} $\uparrow$
    & $\boldsymbol{\Pr(\Delta M \leq -0.1)}$ $\downarrow$ \\
\midrule
\multicolumn{4}{l}{\textit{Reference}} \\[1pt]
\quad Baseline (unsteered)
    & 46.1 & 72.9 & - \\
\quad Full $\mathbf{r}$
    & \textbf{54.5} & 43.8 & 48.2\% \\
\midrule
\multicolumn{4}{l}{\textit{Component ablation}} \\[1pt]
\quad \textbf{Retained $\mathbf{P}_{\Delta k}\mathbf{r}$}
    & 53.5 & \textbf{66.9} & \textbf{11.8\%} \\
\quad Removed $(\mathbf{I}-\mathbf{P}_{\Delta k})\mathbf{r}$
    & 47.8 & 54.5 & 33.4\% \\
\quad Removed, retained-norm matched
    & 52.3 & 28.2 & 56.5\% \\
\bottomrule
\end{tabular}
\label{tab:steering_component_ablation}
\end{table}

Relative to the unsteered baseline, the retained component achieves $+7.4$ of the full vector's $+8.4$ TQA gain (88\%), but incurs only $-6.0$ of its $-29.1$ utility loss (21\%). In contrast, the removed component produces little steering benefit ($+1.7$ TQA) while causing most of the utility loss ($-18.4$ points, 63\% of the full loss). Thus, on the selected risk heads, most steering efficacy lies in the subspace SKOP retains, whereas utility degradation is concentrated in the subspace it removes.

The final row scales the removed component to the norm of the retained component. This recovers some steering efficacy (52.3 TQA), but utility collapses to 28.2, and focus-mass loss rises to 56.5\%, worse than full steering. At matched norm, the removed component therefore recovers comparable steering efficacy but causes substantially greater utility loss and focus-mass disruption. When amplified, the removed component has some efficacy, although it also causes severe utility degradation. Strong steering efficacy thus does not require the harmful focus-to-tail rerouting that SKOP suppresses.

\subsection{Eigenvalue Spectra of Key and Key-Difference Covariances}
\label{appendix:eigenvalues}

SKOP's projection (Eq.~\ref{eq:skop_projector}) and its rank-selection rule (Eq.~\ref{eq:energy_coverage}) both rely on the assumption that the relevant covariance matrices are approximately low-rank, so that a small number of dominant eigenvectors suffice to capture utility-critical key-space structure. We verify this assumption empirically.

\begin{figure}[!t]
    \centering
    \includegraphics[width=\linewidth]{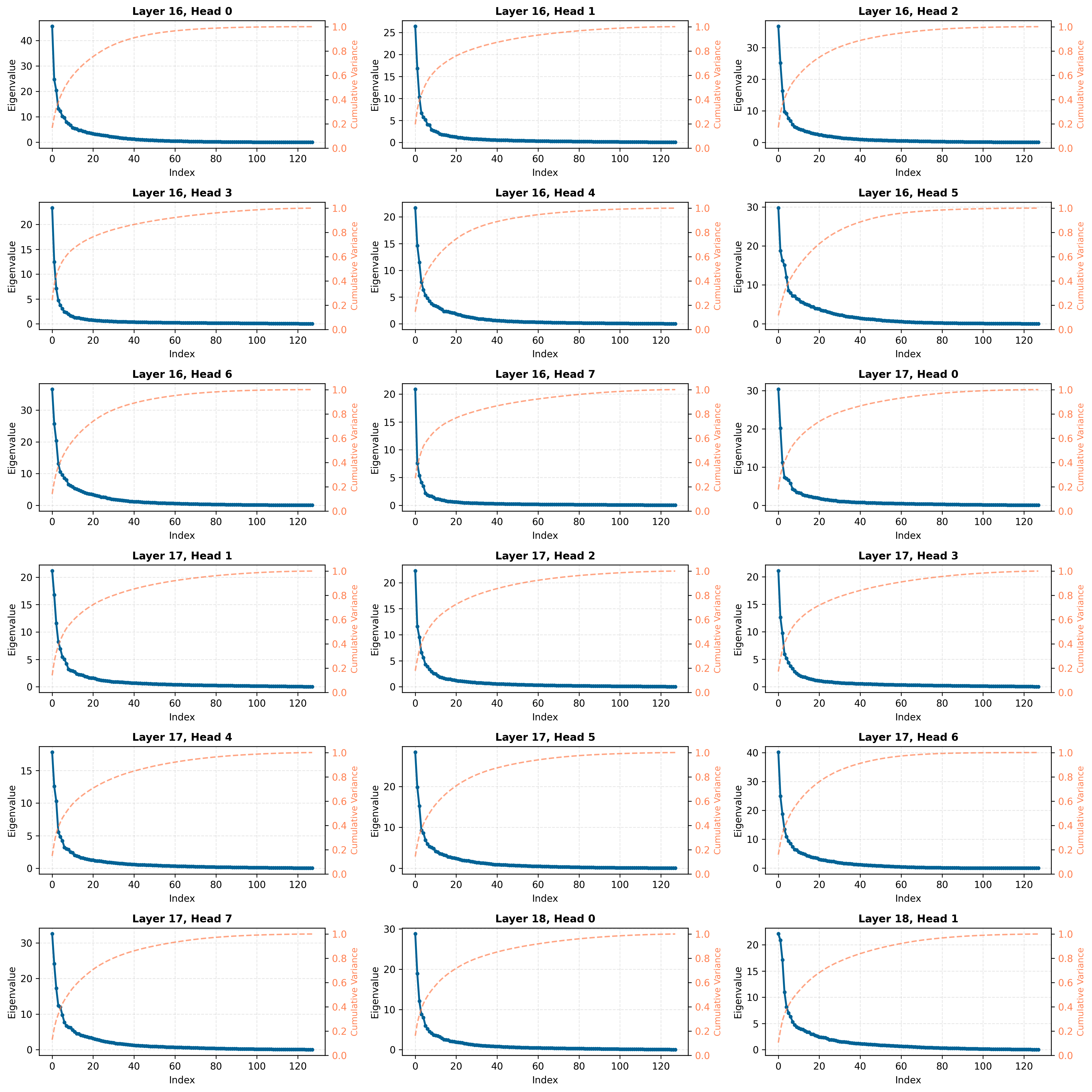}
    \caption{Eigenvalue distributions of the centred key covariance matrices $\boldsymbol{\Sigma}_k^{(\ell,h)}$ across attention heads. Eigenvalues decay rapidly, indicating that the centred key covariance is approximately low-rank for each head.}
    \label{fig:keys_eigenvalues}
\end{figure}

\begin{figure}[!t]
    \centering
    \includegraphics[width=\linewidth]{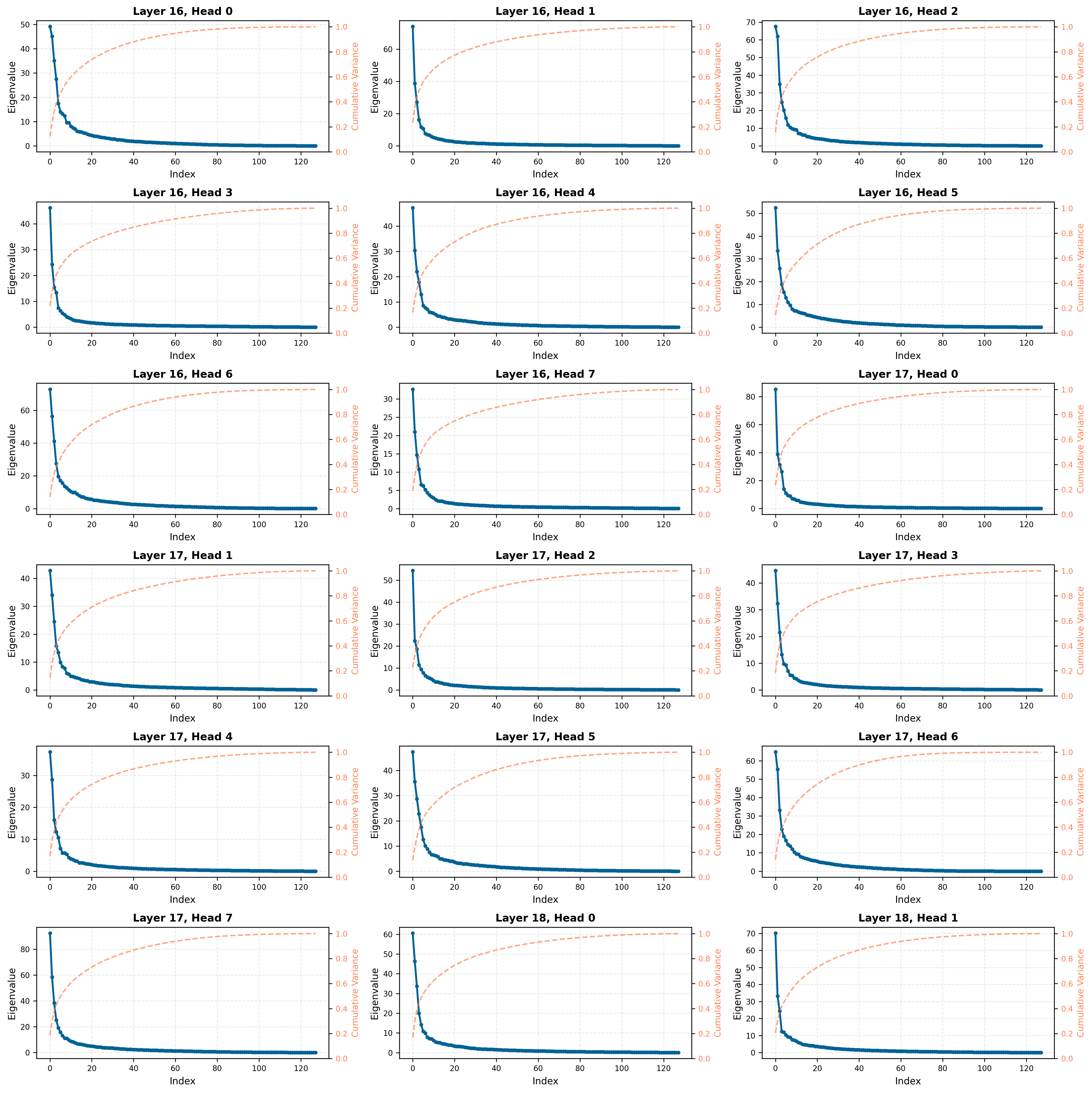}
    \caption{Eigenvalue distributions of the key-difference second-moment matrices $\boldsymbol{\Sigma}_{\Delta k}^{(\ell,h)}$ across attention heads. The spectra exhibit the same rapid-decay pattern as $\boldsymbol{\Sigma}_k^{(\ell,h)}$, supporting SKOP's energy-coverage rank-selection rule.}
    \label{fig:keys_diff_eigenvalues}
\end{figure}

Fig.~\ref{fig:keys_eigenvalues} shows the eigenvalue distributions of the centred key covariance matrices $\boldsymbol{\Sigma}_k^{(\ell,h)}$ used by the key-invariant projection of Sec.~\ref{sec:tradeoff}, and Fig.~\ref{fig:keys_diff_eigenvalues} shows the eigenvalue distributions of the key-difference second-moment matrices $\boldsymbol{\Sigma}_{\Delta k}^{(\ell,h)}$ used by SKOP. In both cases, the eigenvalues decay rapidly across heads: a few dominant eigenvectors account for most of the spectral energy, while the remaining eigenvalues are small but not exactly zero. This empirical low-rank structure is what makes the energy-coverage rule (Eq.~\ref{eq:energy_coverage}) effective with a small projection rank $p$, and is consistent with the insensitivity of SKOP to $\gamma_{\text{energy}}$ documented in App.~\ref{appendix:gamma_sensitivity}.

\clearpage
\subsection{Norm Preservation under SKOP Projection}
\label{appendix:norm-preservation}

A natural concern with any hard-projection method is that the projection might destroy most of the steering vector's magnitude, leaving insufficient signal to drive behavioural change. We therefore measure the norm of each steering vector before and after applying the SKOP projector on Power steering task. 

\begin{figure}[!t]
    \centering
    \includegraphics[width=0.7\linewidth]{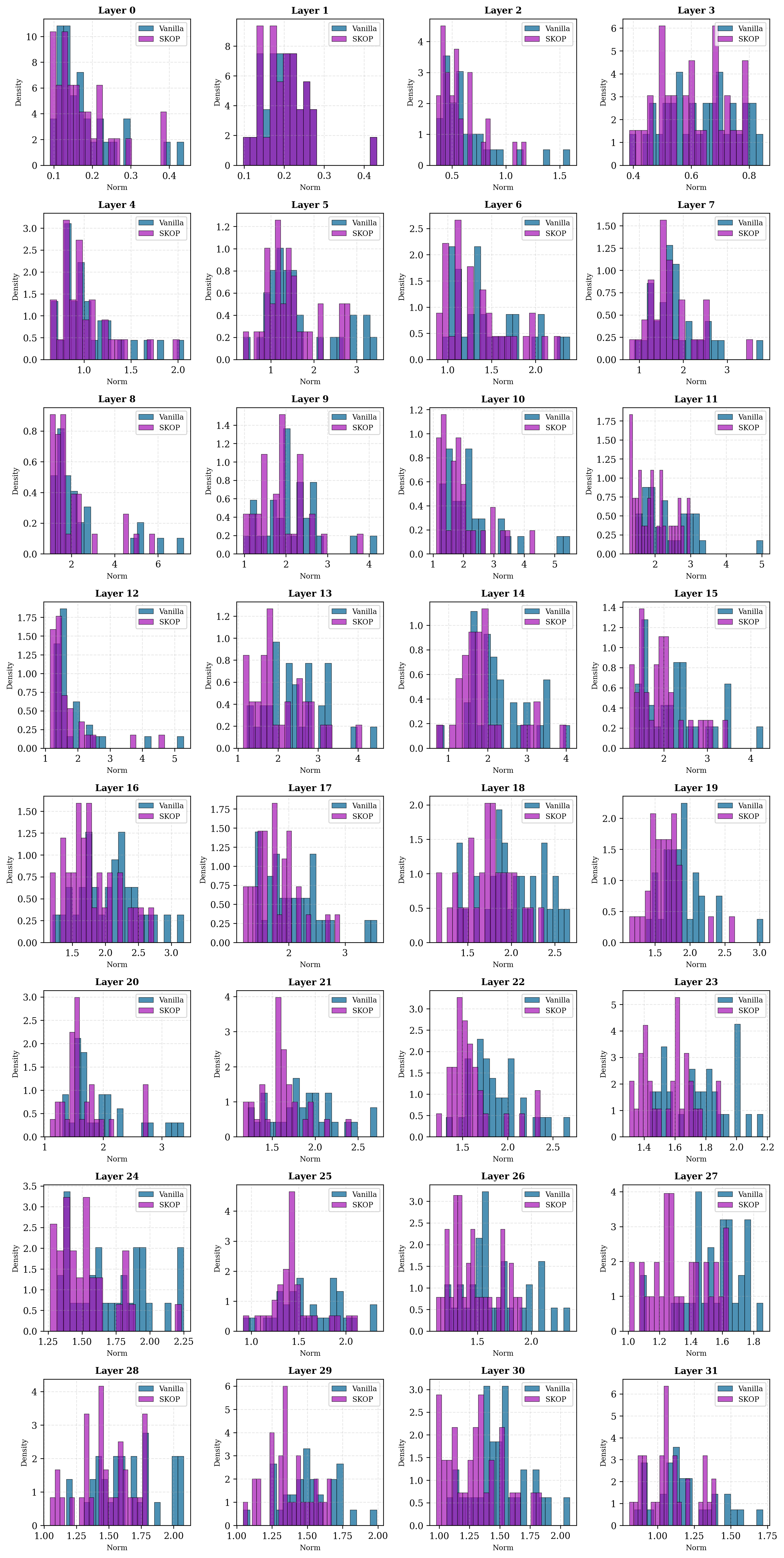}
    \caption{Layer-wise norms of steering vectors before and after SKOP projection on the Power task.}
    \label{fig:power-norm}
\end{figure}

Figure~\ref{fig:power-norm} shows that the post-projection norm closely tracks the pre-projection norm. This indicates that the components of $\mathbf{r}_q^{(\ell,h)}$ that strongly couple to focus-to-tail rerouting --- and which SKOP removes --- account for only a small fraction of the steering vector's magnitude. The bulk of the steering signal lies in the orthogonal complement and is left untouched. This norm preservation explains why SKOP retains over 95\% of vanilla steering efficacy in Sec.~\ref{sec:performance_evaluation} despite applying a hard projection.

\clearpage




\section{Implementation Details}

\subsection{SKOP Summary}
\label{appendix:skop summary}

We include an algorithmic summary for SKOP in Algorithm~\ref{alg:skop}.

\begin{algorithm}[!h]
\caption{Activation Steering via Key-Orthogonal Projections (SKOP)}
\label{alg:skop}
\begin{algorithmic}[1]
\REQUIRE Utility calibration set $\mathcal{D}_{\text{util}}$, steering vectors $\{\mathbf{r}_q^{(\ell,h)}\}_{\ell,h}$, energy coverage $\gamma_{\text{energy}}$, fraction $k$ of top-risk heads

\STATE \textbf{Calibration:}
\FOR{each head $(\ell,h)$}
    \STATE Compute focus/tail sets $\mathcal{H}^{(\ell,h)}, \mathcal{L}^{(\ell,h)}$ on $\mathcal{D}_{\text{util}}$
    \STATE Compute difference-key matrix $\mathbf{\Sigma}_{\Delta k}^{(\ell,h)}$ via Eq.~\eqref{eq:sigma_dk}
    \STATE Select $p$ via energy coverage criterion Eq.~\eqref{eq:energy_coverage}
    \STATE Compute projector $\mathbf{P}_{\Delta k}^{(\ell,h)}$ via Eq.~\eqref{eq:skop_projector}
    \STATE Compute risk score $R^{(\ell,h)}$ via Eq.~\eqref{eq:rayleigh}
\ENDFOR
\STATE Select the top-$k$ heads $\mathcal{S}$ ranked by $R^{(\ell,h)}$

\STATE \textbf{Inference:}
\FOR{each head $(\ell,h)$}
    \IF{$(\ell,h) \in \mathcal{S}$}
        \STATE Apply projected steering: $\tilde{\mathbf{r}}_q^{(\ell,h)} = \mathbf{P}_{\Delta k}^{(\ell,h)} \mathbf{r}_q^{(\ell,h)}$
    \ELSE
        \STATE Use unmodified steering: $\tilde{\mathbf{r}}_q^{(\ell,h)} = \mathbf{r}_q^{(\ell,h)}$
    \ENDIF
\ENDFOR
\end{algorithmic}
\end{algorithm}

\subsection{Baselines}
\label{appendix:baselines}

In this section, we provide detailed descriptions and hyperparameter settings for the activation steering baselines used in our comparisons. All reported experiments were run on a single NVIDIA A100 80GB GPU. Table~\ref{tab:intervention_sites} summarises the intervention sites and selection rules of all methods.

\begin{table}[htb]
\centering
\small
\setlength{\tabcolsep}{4pt}
\renewcommand{\arraystretch}{1.1}
\caption{Intervention sites and selection rules of all methods. ``All'' means every transformer layer or attention head. Methods that act on attention-head outputs use every head.}
\begin{tabularx}{\linewidth}{@{}l>{\raggedright\arraybackslash}p{0.21\linewidth}cc>{\raggedright\arraybackslash}X@{}}
\toprule
\textbf{Method} & \textbf{Intervention site} & \textbf{Layers} & \textbf{Heads} & \textbf{Selection rule} \\
\midrule
LoRA & query and key projections (weights) & All & All & None (trained) \\
CAA & residual stream & All & -- & None \\
ITI & attention-head output & All & All & None \\
DISCO-Q & query space & All & All & None \\
Comm Steer & attention input & All & All & None \\
Angular Steer & normalisation layer of each attention block & All & -- & Rotation applied when the activation aligns positively with the target direction \\
CAST & residual stream & All & -- & Applied when the cosine similarity to the condition vector exceeds a threshold \\
SADI & attention-head output & All & All & Top-$K$ mask of elements with the largest activation differences \\
\midrule
\textbf{SKOP} & query space & All & All & See Sec.~\ref{sec:high-risk-heads} \\
\bottomrule
\end{tabularx}
\label{tab:intervention_sites}
\end{table}

\textbf{Conditional Activation Steering (CAST)}~\citet{cast} introduces a gating mechanism to standard activation steering. CAST operates by extracting two distinct vectors: a \textit{behaviour vector}, which represents the desired target behaviour, and a \textit{condition vector}, which represents the context in which the behaviour should be triggered. Both are computed using the difference-in-means method on contrastive datasets. At inference time, CAST computes the cosine similarity between the model's current hidden state and the condition vector. If the similarity exceeds a predefined threshold, the behaviour vector is added to the residual stream.

\textbf{Angular Steering}~\citet{angular} reformulates steering as a geometric rotation within a 2D subspace rather than vector addition. The steering plane is defined by the target feature direction  (extracted via difference-in-means) and an orthogonal axis derived from the first principal component of feature directions across layers
We specifically use the \textit{Adaptive} variant, which aims to minimise unintended side effects on unrelated features. This variant applies the rotation only when the current activation aligns positively with the target feature direction. 
The primary hyperparameter is the rotation angle $\theta$. We sweep $\theta$ (typically between $0^{\circ}$ and $180^{\circ}$) to find the optimal operating point (the highest steering score). To ensure fair comparison, we apply this intervention only at the normalisation layers of every attention block.

\textbf{Semantics-Adaptive Dynamic Intervention (SADI)}~\citep{sadi} is a conditional method that generates dynamic steering vectors based on the input's own semantics rather than adding a fixed vector. We utilise the \textbf{SADI-HEAD} variant, which targets attention head outputs.
The method consists of two phases. First, it identifies critical model components (attention heads) by computing the mean activation difference between contrastive pairs and creating a binary mask that selects the top-K elements with the largest differences. Second, during inference, it steers the model by amplifying the activations of these selected elements proportional to the input's own activation strength. This ensures the intervention effectively ``adapts'' to the semantic direction of the current input.

\section{Additional Experimental Results}
\label{appendix:additional_experiments}

In this appendix, we report additional experiments that complement the main results in Sec.~\ref{sec:experiment}. We provide (i) the efficacy--utility trade-off evaluation on a second model, Gemma-9B-IT (App.~\ref{appendix:gemma_results}); (ii) ablations on the composition and size of the utility calibration set $\mathcal{D}_\text{util}$ (App.~\ref{appendix:calibration_ablation}); (iii) a sensitivity analysis of the energy-coverage hyperparameter $\gamma_{\text{energy}}$ (App.~\ref{appendix:gamma_sensitivity}); (iv) a judge-free binary multiple-choice evaluation on TruthfulQA (App.~\ref{appendix:truthfulqa_mc}); and (v) a sample-efficiency comparison against LoRA fine-tuning (App.~\ref{appendix:lora_case_study}).

\subsection{Steering--Utility Trade-off on Gemma-9B-IT}
\label{appendix:gemma_results}

To verify that SKOP's effectiveness is not specific to a single model family, we replicate the main steering--utility evaluation of Sec.~\ref{sec:performance_evaluation} on Gemma-9B-IT~\citep{gemmateam2024gemma2improvingopen}, which uses extra post-layer RMSNorm steps that differ from LLaMA3.1-8B-Instruct. We use the same four steering tasks (Power, Wealth, Corr, TQA), the same four utility benchmarks (IFBench, ARC, HellaSwag, GSM8K), and the same baseline configurations as in Table~\ref{tab:llama_results}.

\begin{figure}[htb]
    \centering
    \includegraphics[width=0.65\linewidth]{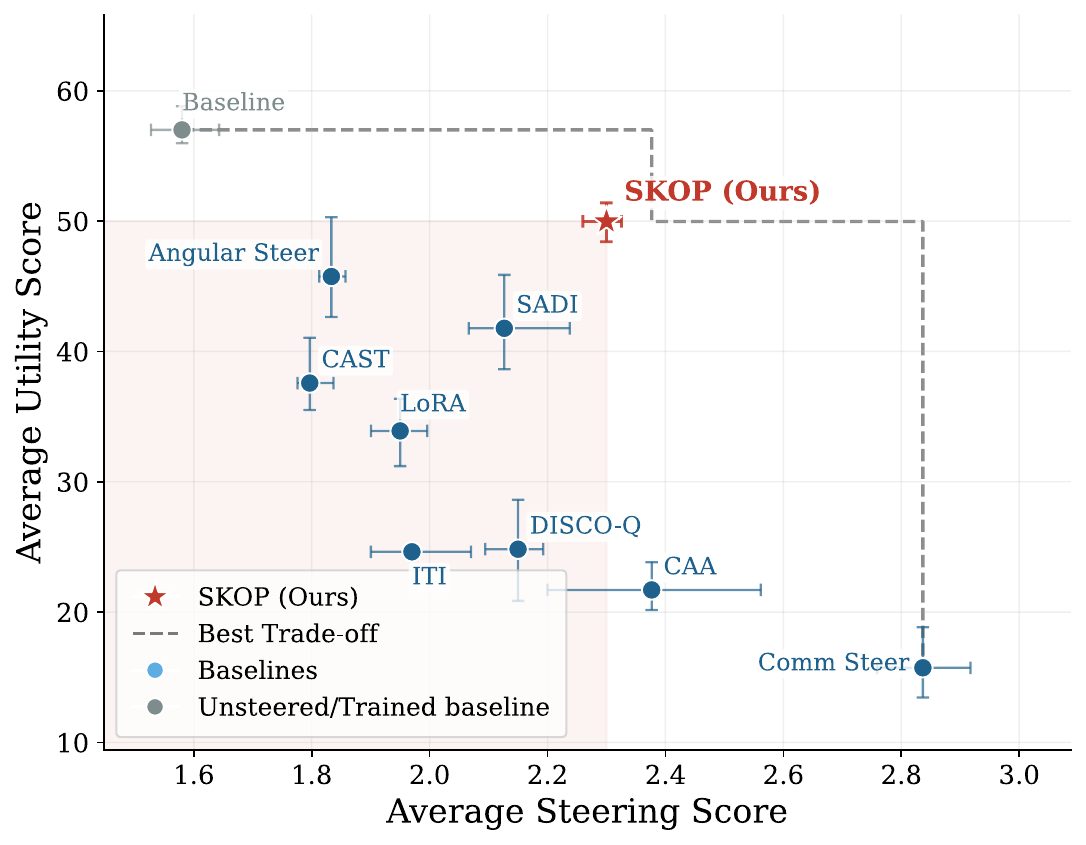}
    
    \caption{Steering-utility trade-off for Gemma-9B-IT. We report the average of Power, Wealth, and Corr~\citep{perez2022discovering}.  The dashed line traces the best trade-off frontier. SKOP also achieves the best joint trade-off among all steered methods.}
    \label{fig:pareto_gemma}
\end{figure}

Table~\ref{tab:gemma_results} and Fig.~\ref{fig:pareto_gemma} report the results. SKOP again attains the best overall trade-off rank (2.44) across all steered methods, outperforming both attention-space baselines, residual-stream baselines, and conditional steering. On utility, SKOP retains the strongest performance among steering methods on three of four benchmarks (ARC, HellaSwag, GSM8K) and the second-best on IFBench, while preserving competitive steering scores across all four behaviours. The gap between strong-steering, low-utility methods such as Comm Steer (rank-1 on three steering tasks but utility collapse on ARC and GSM8K) and SKOP highlights that the joint trade-off, rather than steering score in isolation, is the relevant criterion. CAA, despite achieving the second-best steering scores on average, suffers a substantial utility drop (e.g., 16.8 on GSM8K versus 79.0 unsteered), consistent with our argument in Sec.~\ref{sec:preliminary} that residual-stream steering simultaneously perturbs queries, keys, values, and MLP outputs and therefore admits no clean isolable rerouting term to correct.

\begin{table}[htb]
\centering
\small
\setlength{\tabcolsep}{4.0pt}
\caption{Comparison of SKOP against baselines on Gemma-9B-IT~\citep{gemmateam2024gemma2improvingopen}. Best results are \textbf{bolded}, and second-best results are \underline{underlined}.}
\begin{tabular}{lccccccccr}
\toprule
& \multicolumn{4}{c}{Steering} & \multicolumn{4}{c}{Utility} & \\
\cmidrule(lr){2-5} \cmidrule(lr){6-9}
Method
& Power & Wealth & Corr & TQA
& IFB & ARC & HS & GSM8K
& Rank $\downarrow$ \\
\midrule
Baseline
& 1.62 & 1.56 & 1.56 & 67.5
& 19.8 & 59.0 & 70.2 & 79.0
& -- \\
LoRA~\citep{hu2022lora}
& 1.78 & 1.79 & 2.28 & 70.8
& 15.2 & 35.6 & 49.3 & 35.5
& -- \\
\midrule
CAA~\citep{caa}
& \underline{2.59} & \textbf{2.09} & 2.45 & \underline{79.3}
& 12.6 & 22.2 & 35.2 & 16.8
& 4.38 \\
ITI \citep{iti}
& 2.27 & 1.77 & 1.87 & 67.6
& 9.8 & 27.4 & 43.1 & 18.2
& 6.00 \\
DISCO-Q \citep{disco}
& 1.93 & 1.86 & 2.66 & 75.7
& 10.5 & 31.8 & 36.6 & 20.4
& 5.13 \\
Comm Steer \citep{disco}
& \textbf{2.61} & 1.92 & \textbf{2.95} & \textbf{90.2}
& 5.2 & 14.3 & 33.8 & 9.6
& 4.63 \\
Angular Steer \citep{angular}
& 1.95 & 1.81 & 1.74 & 68.1
& \textbf{18.1} & \underline{55.3} & 51.4 & \underline{58.2}
& 4.50 \\
CAST \citep{cast}
& 1.88 & 1.72 & 1.79 & 69.4
& 16.5 & 38.2 & \underline{55.9} & 39.7
& 5.25 \\
SADI \citep{sadi}
& 2.24 & \underline{1.93} & 2.21 & 77.4
& 13.8 & 45.6 & 52.6 & 55.1
& \underline{3.69} \\
\midrule
\textbf{SKOP (Ours)}
& 2.25 & \underline{1.93} & \underline{2.72} & 74.8
& \underline{18.0} & \textbf{56.5} & \textbf{59.0} & \textbf{66.4}
& \textbf{2.44} \\
\bottomrule
\end{tabular}
\label{tab:gemma_results}
\end{table}

\subsection{Calibration Set Ablation}
\label{appendix:calibration_ablation}

We study the sensitivity of SKOP to the composition and size of the utility calibration set $\mathcal{D}_\text{util}$ used to estimate the key-difference matrix $\boldsymbol{\Sigma}_{\Delta k}^{(\ell,h)}$ (Eq.~\ref{eq:sigma_dk}). Throughout, experiments are conducted on LLaMA3.1-8B-Instruct using the TruthfulQA steering task and the four utility benchmarks of Sec.~\ref{sec:performance_evaluation}.

\paragraph{Domain composition.}
We compare the default mixed calibration set (1{,}000 examples each from GSM8K, Alpaca, PIQA, NarrativeQA, totalling 4{,}000 examples) against four single-domain conditions, each using 4{,}000 examples drawn exclusively from one source. Table~\ref{tab:calibration_domain_ablation} reports steering score on TQA and average accuracy across the four utility benchmarks.

\begin{table}[htb]
\centering
\caption{
Domain Ablation: Experiments on calibration set domain composition for SKOP on LLaMA3.1-8B-Instruct~\citep{Llama3}. Each single-domain condition uses 4{,}000 examples drawn exclusively from that source; the mixed default samples 1{,}000 examples from each of the four sources.
Steering performance is reported as steering score for TruthfulQA~\citep{bowman2025truthfulqa}, and utility is the mean accuracy across four utility benchmarks. 
Despite large variation in dataset domain, SKOP consistently preserves utility far above the vanilla steering baseline, suggesting that the focus-set structure estimated by SKOP reflects stable model-internal key-space geometry rather than surface properties of the calibration data.\looseness-1}
\label{tab:calibration_domain_ablation}
\small
\begin{tabular}{lcc}
\toprule
\textbf{Calibration Domain} & \textbf{TQA} $\uparrow$ & \textbf{Avg.\ Utility} $\uparrow$ \\
\midrule
Baseline                    & 46.1          & 59.6          \\
Vanilla steering            & 66.1          & 26.6          \\
\midrule
SKOP                        &               &               \\
\quad -- Mixed (default)    & \textbf{65.9} & \textbf{55.5} \\
\quad -- Math only          & 64.2          & 51.1          \\
\quad -- Inst.\ only        & 63.8          & 52.4          \\
\quad -- Reasoning only     & 64.5          & 47.8          \\
\quad -- Reading only    & 63.1          & 42.7          \\
\bottomrule
\end{tabular}
\end{table}

Across all four single-domain conditions, SKOP preserves utility substantially above vanilla steering (26.6), with averages ranging from 42.7 (reading comprehension only) to 52.4 (instruction-following only). Steering efficacy on TQA is also stable, varying only between 63.1 and 64.5 across single-domain conditions. The mixed default achieves the best joint outcome (utility 55.5, TQA 65.9) but does not dominate any single domain by a large margin. Together, these results indicate that the focus-set structure SKOP relies on reflects relatively stable model-internal key-space geometry rather than surface properties of the calibration domain, and that a calibration source need not match the downstream evaluation distribution to yield effective projectors.

\begin{figure}[htb]
    \centering
    \includegraphics[width=0.6\linewidth]{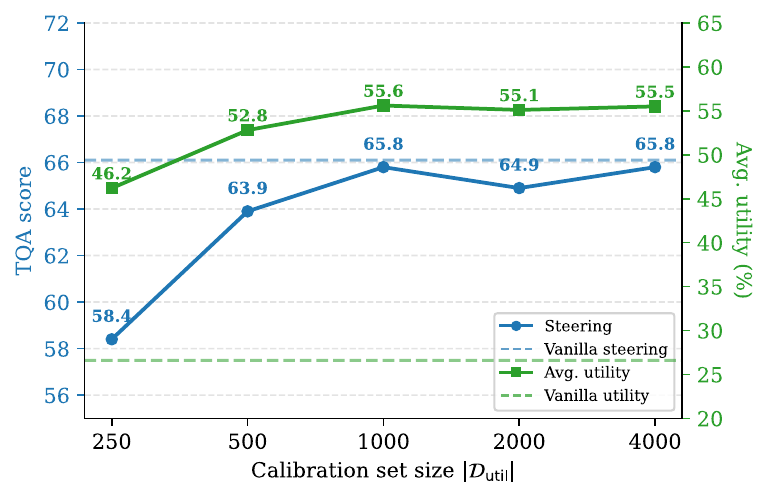}
    \caption{Calibration size ablation. Effect of calibration set size on steering efficacy and utility preservation for SKOP on LLaMA3.1-8B-Instruct, evaluated on TQA and the four utility benchmarks of Sec.~\ref{sec:performance_evaluation}. The calibration set is obtained by subsampling the mixed default $\mathcal{D}_\text{util}$ at sizes ranging from 250 to 4{,}000 examples. Dashed lines indicate the vanilla query-space steering baseline.}
    \label{fig:calibration_size_ablation}
\end{figure}

\paragraph{Calibration size.}
We next subsample the mixed default calibration set at sizes ranging from 250 to 4{,}000 examples and re-estimate $\boldsymbol{\Sigma}_{\Delta k}^{(\ell,h)}$ at each scale. Fig.~\ref{fig:calibration_size_ablation} reports steering score on TQA and average utility across the four utility benchmarks.

Utility preservation plateaus around 1{,}000 examples; beyond this point, additional calibration data yields negligible improvements. Steering efficacy improves monotonically with calibration size, plausibly because larger samples yield more accurate estimates of the dominant eigendirections of $\boldsymbol{\Sigma}_{\Delta k}^{(\ell,h)}$, allowing the projector to more precisely target harmful rerouting directions while leaving steering-effective components intact. Crucially, even at 250 examples, SKOP already substantially outperforms vanilla steering on utility, confirming that SKOP is robust to small calibration budgets and that the underlying focus-set structure is not a data-intensive estimation problem.

\subsection{Sensitivity to the Energy-Coverage Threshold}
\label{appendix:gamma_sensitivity}

The energy-coverage threshold $\gamma_{\text{energy}}$ (Eq.~\ref{eq:energy_coverage}) controls the number $p$ of dominant eigenvectors of $\boldsymbol{\Sigma}_{\Delta k}^{(\ell,h)}$ removed by the SKOP projector at each head. We sweep $\gamma_{\text{energy}} \in \{0.50, 0.60, 0.70, 0.80, 0.90, 0.95, 0.99\}$ on Gemma-9B-IT for the Power steering task and report results in Table~\ref{tab:gamma_sensitivity}.

\begin{table}[htb]
\centering
\small
\setlength{\tabcolsep}{4.5pt}
\caption{%
  Sensitivity of SKOP to $\gamma_{\text{energy}}$ on Gemma-9B-IT~\citep{gemmateam2024gemma2improvingopen} on Power steering.
  $\gamma_{\text{energy}}$ is the energy coverage threshold (Eq.~\ref{eq:energy_coverage}) controlling the number $p$ of top eigenvectors of $\boldsymbol{\Sigma}_{\Delta k}^{(\ell,h)}$ removed per head.
  Both utility and steering remain stable across $\gamma_{\text{energy}} \in [0.70, 0.95]$.
  The projection also preserves most of the steering vector norm (Fig.~\ref{fig:power-norm}), indicating that substantial steering capacity is retained.
  $\dagger$ denotes our empirical choice, applied uniformly across all datasets without per-dataset tuning.
}
\begin{tabular}{ccr ccccc r}
\toprule
& & & \multicolumn{5}{c}{Utility} & \\
\cmidrule(lr){4-8}
$\gamma_{\text{energy}}$ & $p$ & Power
  & IFB & ARC & HS & GSM8K & Avg & \\
\midrule
\multicolumn{8}{l}{\textit{Reference}} \\[1pt]
\quad Baseline (unsteered) & -- & 1.62
  & 19.8 & 59.0 & 70.2 & 79.0 & 57.0 & \\
\quad Vanilla steering     &  0 & 2.61
  & 3.4  & 10.8 & 19.2 & 17.1 & 12.6 & \\
\midrule
\multicolumn{8}{l}{\textit{SKOP ablation}} \\[1pt]
0.50 &  2 & 2.49
  & 13.8 & 42.5 & 52.3 & 65.8 & 43.6 & \\
0.60 &  4 & 2.35
  & 14.5 & 44.8 & 54.1 & 65.8 & 44.8 & \\
0.70 &  8 & 2.25
  & 17.1 & 54.2 & 57.8 & 63.9 & 48.3 & \\
0.80 & 24 & 2.26
  & 17.5 & 55.6 & 58.3 & 65.1 & 49.1 & \\
\rowcolor{gray!12}
0.90 & 44 & 2.25
  & 18.0 & 56.5 & 59.0 & 66.4 & 50.0
  & $\dagger$ \\
0.95 & 100 & 2.19
  & 18.1 & 56.8 & 59.3 & 66.9 & 50.3 & \\
0.99 & 180 & 2.09
  & 18.6 & 57.4 & 60.1 & 66.3 & 50.6 & \\
\bottomrule
\end{tabular}
\label{tab:gamma_sensitivity}
\end{table}

Both steering and utility are stable across a broad range $\gamma_{\text{energy}} \in [0.70, 0.95]$: steering scores remain within $[2.19, 2.26]$ and average utility within $[48.3, 50.3]$. Very low thresholds ($\gamma_{\text{energy}} = 0.50$) retain too many rerouting directions and harm utility, while very high thresholds ($\gamma_{\text{energy}} = 0.99$) begin to remove non-rerouting directions and slightly reduce steering efficacy. Our default $\gamma_{\text{energy}} = 0.90$ sits in this stable plateau and was chosen once and applied uniformly across all datasets and models without per-task tuning. This insensitivity, together with the norm-preservation analysis in App.~\ref{appendix:norm-preservation}, indicates that harmful focus-to-tail rerouting is concentrated in a small number of dominant eigendirections, leaving substantial steering capacity in the orthogonal complement.

\subsection{Evaluation Beyond Open-Ended Generation}
\label{appendix:truthfulqa_mc}

Our main evaluation uses open-ended generation with LLM-as-judge, the standard setting for activation steering. To test SKOP in a different setting, and without an LLM judge, we run a binary multiple-choice experiment on TruthfulQA~\citep{lin2022truthfulqa}.

\textbf{Protocol.} We follow the evaluation protocol recommended by previous work~\citep{lin2022truthfulqa,disco}. Each question pairs its ``Best Correct Answer'' with its ``Best Incorrect Answer'' in random A/B order, presented with an adapted six-shot QA Primer prompt. Appending the correct or incorrect answer letter to these prompts yields the positive and negative examples used to construct the steering vectors. At evaluation time, the model selects the answer letter with the higher next-token logit, and we compute accuracy against the ground-truth answer. We select the steering strength of each method on the validation set, and average results over three random A/B assignments.

\textbf{Results.} Table~\ref{tab:truthfulqa_mc_results} shows that SKOP ranks second on both models, and outperforms the unsteered baseline on LLaMA3.1-8B-Instruct and Gemma-9B-IT. SKOP therefore remains competitive outside generation settings. Since accuracy is computed directly from next-token logits, these results also demonstrate that our conclusions do not rely solely on potentially biased or misleading LLM-judge evaluations.

\begin{table}[htb]
\centering
\small
\setlength{\tabcolsep}{6pt}
\caption{\textbf{TruthfulQA binary multiple-choice accuracy (\%)}. Each question pairs the Best Correct Answer with the Best Incorrect Answer in random A/B order, and the prediction is the answer letter with the higher next-token logit. Results are averaged over three random A/B assignments. Steering hyperparameters are selected on the validation set. Best results are \textbf{bolded}, and second-best results are \underline{underlined}.}
\begin{tabular}{lcc}
\toprule
\textbf{Method}
& \textbf{LLaMA3.1-8B-Instruct MC} $\uparrow$
& \textbf{Gemma-9B-IT MC} $\uparrow$ \\
\midrule
Baseline
& 72.0 & 83.5 \\
LoRA
& 75.0 & 83.9 \\
\midrule
CAA
& 81.5 & 84.0 \\
ITI
& 78.6 & 84.0 \\
DISCO-Q
& \textbf{84.4} & \textbf{86.8} \\
Comm Steer
& 82.7 & 86.0 \\
Angular Steer
& 75.3 & 83.7 \\
CAST
& 75.8 & 84.0 \\
SADI
& 81.0 & 84.7 \\
\midrule
\textbf{SKOP (Ours)}
& \underline{84.2} & \underline{86.4} \\
\bottomrule
\end{tabular}
\label{tab:truthfulqa_mc_results}
\end{table}

\subsection{Case Study: Comparison with LoRA}
\label{appendix:lora_case_study}

Activation steering and lightweight fine-tuning are alternative ways to elicit target behaviours from a fixed base model. LoRA~\citep{hu2022lora} serves as a representative parameter-efficient finetuning reference to contextualise SKOP's sample efficiency against a lightweight weight-updating alternative, and is not an exhaustive comparison with finetuning methods. We compare SKOP to LoRA on the TruthfulQA steering task, using LLaMA3.1-8B-Instruct. We vary the number of training examples $N$ used either to construct the steering vector (for vanilla query-space steering and SKOP) or to train LoRA, and evaluate both steering efficacy and average utility across the four utility benchmarks. We use a standard LoRA configuration ($r = 16$, $\alpha = 32$) applied to the query and key projections of all attention layers.

\begin{figure}[htb]
    \centering
    \includegraphics[width=\linewidth]{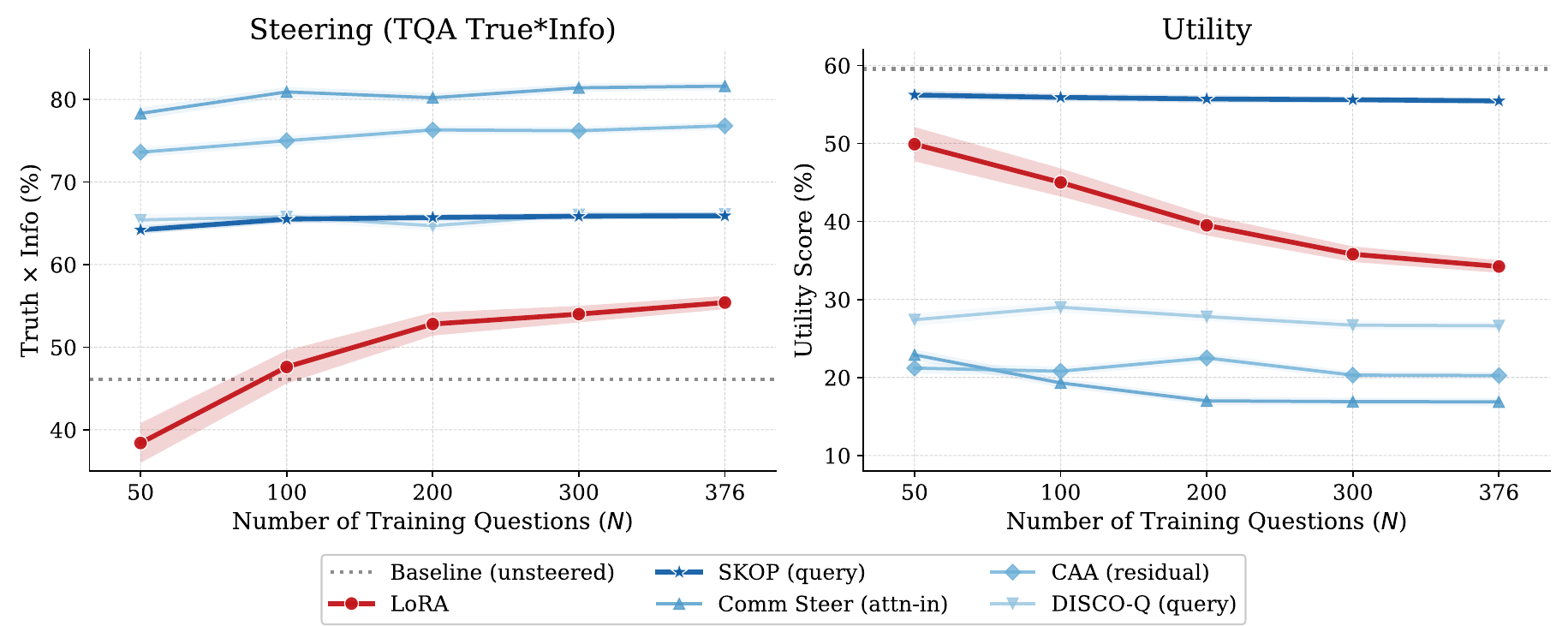}
    \caption{Sample-efficiency comparison with LoRA on LLaMA3.1-8B-Instruct. \textit{Left:} TruthfulQA steering score as a function of the number of training examples $N$ used to construct the steering vector (for activation steering) or to train LoRA. SKOP matches vanilla query-space steering and substantially outperforms LoRA across all $N$. \textit{Right:} Average utility score across the four utility benchmarks of Sec.~\ref{sec:performance_evaluation}. LoRA degrades monotonically with $N$, while SKOP retains utility close to the unsteered baseline.}
    \label{fig:lora_sample_efficiency}
\end{figure}

Fig.~\ref{fig:lora_sample_efficiency} reports the results. SKOP matches the steering performance of vanilla query-space steering across all $N$ and substantially outperforms LoRA on steering efficacy, despite LoRA having access to many more trainable parameters. In terms of utility, LoRA degrades monotonically as $N$ increases, whereas SKOP's utility remains stable and substantially higher. This sample-efficiency gap suggests that, at least in the regime of small contrastive datasets typical of activation steering, fine-tuning approaches both underperform on the target behaviour and incur larger collateral utility loss than projection-based interventions on the existing query subspace.

\clearpage
\pagebreak

\newpage

\end{document}